\documentclass[lettersize,journal]{IEEEtran}
\usepackage[utf8]{inputenc}
\usepackage{amsthm}
\usepackage{import}
\usepackage{algorithm,algorithmic}
\usepackage{lipsum}
\usepackage{color} 
\usepackage{multirow}
\usepackage{graphicx}
\usepackage{float}
\usepackage{subfigure}
\usepackage{booktabs}
\usepackage{multicol}
\usepackage[nosort]{cite}
\usepackage{multirow}
\usepackage[normalem]{ulem}
\usepackage{adjustbox}
\usepackage{emptypage}
\usepackage{amsfonts}
\usepackage{amssymb}
\usepackage[figuresright]{rotating}
\usepackage{hhline}
\usepackage{diagbox}
\usepackage{booktabs}
\usepackage{makecell}
\usepackage{subfigure} 
\usepackage{amsmath}
\usepackage{mathrsfs}

\renewcommand\tabcolsep{5pt}

\usepackage[normalem]{ulem}
\useunder{\uline}{\ul}{}

\theoremstyle{plain}
\newtheorem{definition}{Definition}

\theoremstyle{plain}

\newtheorem{theorem}{Theorem}

\begin{document}

\title{Bayesian Robust Tensor Ring Decomposition for Incomplete Multiway Data}

\author{Zhenhao~Huang, Yuning~Qiu, Xinqi~Chen, Weijun~Sun and Guoxu~Zhou
    \thanks{This work is supported in part by National Natural Science Foundation of China under Grant 62073087, Grant 62071132, Grant 61973090 and Grant 62203124. (\emph{Corresponding authors: Guoxu Zhou.})}
\thanks{The authors are with the School of Automation, Guangdong University of Technology, Guangzhou 510006, China. Z. Huang is also with the Center for Advanced Intelligence Project (AIP), RIKEN, Tokyo, 103-0027, Japan. G. Zhou is also with Key Laboratory of Intelligent Detection and The Internet of Things in Manufacturing, Ministry of Education. (e-mail: zhhuang.gdut@qq.com, yuning.qiu.gd@gmail.com, 1111904005@mail2.gdut.edu.cn, gdutswj@gdut.edu.cn,   gx.zhou@gdut.edu.cn).}
\thanks{This work has been submitted to the IEEE for possible publication. Copyright may be transferred without notice, after which this version may no longer be accessible.}}

\markboth{Journal of \LaTeX\ Class Files,~Vol.~, No.~, ~}%
{Huang \MakeLowercase{\textit{et al.}}: Bayesian Robust Tensor Ring Decomposition for Incomplete Multiway Data}


\maketitle

\begin{abstract}
Robust tensor completion (RTC) aims to recover a low-rank tensor from its incomplete observation with outlier corruption. The recently proposed tensor ring (TR) model has demonstrated superiority in solving the RTC problem. However, the existing methods either require a pre-assigned TR rank or aggressively pursue the minimum TR rank, thereby often leading to biased solutions in the presence of noise. In this paper, a Bayesian robust tensor ring decomposition (BRTR) method is proposed to give more accurate solutions to the RTC problem, which can avoid exquisite selection of the TR rank and penalty parameters. A variational Bayesian (VB) algorithm is developed to infer the probability distribution of posteriors. During the learning process, BRTR can prune off slices of core tensor with marginal components, resulting in automatic TR rank detection. Extensive experiments show that BRTR can achieve significantly improved performance than other state-of-the-art methods. 
\end{abstract}


\section{Introduction}
\label{intro}
Many data in real-world applications are naturally represented as tensors, also known as multi-dimensional arrays. For instance, a color image forms a third-order tensor with the size of width $\times$ height $\times$ channel, and a color video clip forms a fourth-order one with an additional dimension of time frame. Tensor data-based learning methods, including CANDECOMP/PARAFAC (CP) decomposition \cite{qiu2021canonical} and Tucker decomposition \cite{zhou2015efficient, huang2022dynamic, qiu2020generalized, qiu2021approximately, qiu2021semi}, have significant advantages of preserving multi-linear structure information of tensor data \cite{kolda2009tensor, qiu2022efficient}, and have been successfully applied to data mining, clustering and classification problem \cite{he2017pattern, sun2021new} and brain signal processing \cite{huang2021recognition}. 

Most of the time, the observed data are incomplete due to the failure of the data collection tools and other abnormal conditions. Therefore, tensor completion (TC), which aims to predict missing entries from partial observations, is proposed to handle this problem and has been applied to many fields, such as color image and video recovery \cite{liu2019low, zhang2022effective}, knowledge graph completion \cite{shao2022tucker}, feature selection \cite{kuang2018feature, wu2021latent} and link prediction \cite{nickel2013tensor, ermics2015link}. In the meanwhile, the CP model was firstly considered for tensor completion due to its simplicity, such as CP weighted optimization (CP-WORT) \cite{acar2011scalable}, CP nonlinear least squares (CPNLS) \cite{sorber2013optimization}. Moreover, to avoid manual rank selection, Zhao $et\ al$. introduced a Bayesian approach for the CP model where CP rank can be automatically determined. Nevertheless, the CP model has some fundamental limitations, including that the CP decomposition of a given tensor can be ill-posed, and its computation often suffers from serious convergence issues \cite{de2008tensor}. Hence, tensor completion using the Tucker model gained growing interests, such as Tucker weighted optimization (Tucker-WORT) \cite{filipovic2015tucker}, Tucker-based overlapped nuclear norm optimization \cite{liu2012tensor} and scalable Tucker factorization method \cite{oh2018scalable, lee2018fast}. No matter which model is employed, the associated low-rank feature is the key to achieving satisfying recovery performance. The Tucker model-based methods are often accused that the core tensor grows exponentially and is inefficient to deal with high-order tensors. Recently, some new definitions of tensor rank have been proposed to solve TC problem, including tensor tubal rank \cite{zhang2014novel,lu2019low, jiang2020multi, wang2020TNN, wang2020OITNN}, tensor train (TT) rank \cite{oseledets2011tensor} and tensor ring (TR) rank \cite{zhao2016tensor}. Among them, TT and TR models have achieved more satisfactory performance \cite{xu2020learning, huang2020provable,chen2020accommodating, yu2021fast, qiu9800181, liu2021simulated, zhang2021multiscale, yu2020low, yu2021robust}. For TT model, the original tensor is approximated by the tensor contraction of a series of three-order tensors (the first and the last one are matrices). Hence, TT is more suitable for higher-order tensors and a lot of TC methods using TT model were proposed, such as simple low-rank TC via TT (SiLRTC-TT), parallel matrix factorization via TT (TMac-TT) \cite{bengua2017efficient} and matrix factorization based on TT rank and total variation (MF-TTTV) \cite{ding2019low}. As analyzed in \cite{zhao2016tensor}, TT rank has a strict rank requirement on the border factors, resulting in an unbalanced rank scheme and constrained representation ability. TR rank relaxes the strict restriction over TT rank and is able to give a more balanced representation, hence is more suitable for arbitrary high-order tensors and can better capture the low-rank nature of data. Based on TR model, Yuan $et\ al.$ proposed a TR low-rank factors (TRLRF) method by minimizing the nuclear norm of all factor tensors \cite{yuan2019tensor}. Yu $et\ al.$ defined a new circular unfolding operation and proposed a TR nuclear norm minimization (TRNNM) \cite{yu2019tensor}. Long $et\ al.$ proposed a Bayesian low-rank tensor ring method \cite{long2021bayesian} for the TC problem. These works of literature show that the TC methods based on the TR model often have better recovery performance than other methods.

All of the above methods perform well if the partially observed data are free from noise corruption. However, the observed data may be corrupted by noise and outliers in the acquisition conditions and transmission process. It is of great interest to predict the intrinsic tensor from noisy and missing observations, which is also known as the robust tensor completion (RTC) problem. Although RTC is more practical and valuable, there is limited work on it. In \cite{zhao2015bayesian}, Zhao $et\ al.$ proposed a Bayesian robust tensor factorization (BRTF) method, which achieved a significant improvement in processing noisy data under Bayesian treatment. But it still suffers from the same issues as the CP model. Song $et\ al.$ solved RTC problem via transformed tubal nuclear norm (TTNN) \cite{song2020robust}. TTNN used a new unitary transformation that had a better recovery performance compared with the use of the Fourier transformation. But it can only serve 3-order tensor data and requires a hyperparameter to be manually adjusted. Huang $et\ al.$ proposed a robust low-rank TR completion (RTRC) method with recovery guarantee \cite{huang2020robust}. However, RTRC has to unfold the tensor into a matrix, which breaks the structure of the tensor TR decomposition, and it also requires manual adjustment of the parameters. Furthermore, rank minimization model-based algorithms, including TTNN and RTRC, have been proven to deviate from the solution of the original problem in the presence of noise \cite{chartrand2007exact, nie2012low, liu2014exact, mohan2012iterative}.

In this paper, we propose a Bayesian robust tensor ring decomposition (BRTR) method for RTC problem under a fully Bayesian treatment. The BRTR method is modeled by multiplicate interactions of core tensors and an additive sparse noise tensor, where sparsity-inducing priors are placed over them. In addition, we develop a variational Bayesian (VB) algorithm for posterior inference. During the learning process, BRTR can prune off slices of core tensor with marginal components, resulting in automatic TR rank detection. The main advantage is that BRTR can automatically learn TR rank in recovering noisy and missing data. In addition, the low-rank and sparse components can be accurately separated without tuning hyperparameters. Extensive experiments show that BRTR can achieve significantly improved performance than other state-of-the-art methods.

The rest of this paper is organized as follows. Section \ref{preli} clarifies some basic notations and definitions for TR decomposition. In Section \ref{BRTRD}, we introduce the BRTR method under the Bayesian approach and give a VB algorithm for model learning. In Section \ref{exper}, we perform extensive experiments to evaluate our proposed BRTR method in competition with other methods. Finally, the conclusion and future work are drawn in Section \ref{concl}. 
\section{Notations and Preliminaries}
\label{preli}
\subsection{Notations}
In this section, we introduce some basic definitions and their description, which are summarized in TABLE \ref{basic_notation}.
In special, we introduce some concepts about probability theory. $p(\cdot)$ and $q(\cdot)$ denote the probability distributions. $\mathcal{N}(\mu,\sigma^2)$ denotes a Gaussian distribution with mean $\mu$ and variance $\sigma^2$. $\mathrm{Ga}(x|a,b) = (b^{a} x^{a-1} e^{-b x})/{\Gamma(a)}$ denotes a Gamma distribution where ${\Gamma(a)}$ is a gamma function.
\begin{definition}[Matrix Normal Distribution]
	\emph{Given a matrix $\mathbf{A} \in \mathbb{R}^{I \times R}$, its matrix normal distribution is defined as $\mathbf{A} \sim \mathcal{MN} (\tilde{\mathbf{A}},\mathbf{U},\mathbf{V})$,which can be further expressed by}
\end{definition}
\begin{equation}
	\begin{aligned}
		&p(\mathbf{A}\mid \tilde{\mathbf{A}},\mathbf{U},\mathbf{V})\\
		&=\frac{\exp \left(-\frac{1}{2} \operatorname{Tr}\left(\mathbf{V}^{-1}(\mathbf{A}-\tilde{\mathbf{A}})^{\top} \mathbf{U}^{-1}(\mathbf{A}-\tilde{\mathbf{A}})\right)\right)}{(2 \pi)^{I R / 2}|\mathbf{V}|^{I / 2}|\mathbf{U}|^{R / 2}}
	\end{aligned}
\end{equation}
where $\tilde{\mathbf{A}}$ denotes the mean, $\mathbf{U}$ and $\mathbf{V}$ denote diagonal covariance matrices. And we have $\operatorname{vec}({\mathbf{A}}) \sim \mathcal{N}(\operatorname{vec}({\tilde{\mathbf{A}}}), \mathbf{U} \otimes \mathbf{V})$.
\begin{table}[t]
	\centering
	\caption{Some basic definitions and their description.}
	\resizebox{\linewidth}{!}{
		\begin{tabular}{cl}
			\toprule Notation & Description \\
			\hline
			a, $\mathbf{a}$, $\mathbf{A}$, $\boldsymbol{\boldsymbol{\mathcal{A}}}$  & A scalar, vector, matrix, tensor, respectively\\
			$\boldsymbol{\mathcal{A}}_{i_1 \cdots i_N}$ & The ($i_1, \cdots, i_N$) entry of tensor $\boldsymbol{\mathcal{A}}$\\
			$\operatorname{vec}(\boldsymbol{\mathcal{A}})$ & The vectorization of tensor $\boldsymbol{\mathcal{A}}$\\
			$\operatorname{Tr}(\mathbf{A})$ & The trace operation of square matrix $\mathbf{A}$\\
			$\|\boldsymbol{\mathcal{A}}\|_{F}$ & The Frobenius norm of tensor $\boldsymbol{\mathcal{A}}$\\ 
			$<\boldsymbol{\mathcal{A}},\boldsymbol{\mathcal{B}}>$ & The tensor inner product of two tensors $\boldsymbol{\mathcal{A}}$ and $\boldsymbol{\mathcal{B}}$\\ 
			$\boldsymbol{\mathcal{A}} \circledast \boldsymbol{\mathcal{B}}$ & The Hadamard product of two tensors $\boldsymbol{\mathcal{A}}$ and $\boldsymbol{\mathcal{B}}$ \\
			$\boldsymbol{\mathcal{A}} \otimes \boldsymbol{\mathcal{B}} $ & The Kronecker product of two tensors $\boldsymbol{\mathcal{A}}$ and $\boldsymbol{\mathcal{B}}$\\
			\bottomrule
	\end{tabular}}
	\label{basic_notation}
\end{table}
\subsection{Tensor Ring Decomposition}
Given an $N$-order tensor $\boldsymbol{\mathcal{L}} \in \mathbb{R}^{I_1 \times I_2 \times \cdots \times I_N}$, the TR model decomposes it into a sequence of latent tensors $\boldsymbol{\mathcal{Z}}^{(n)} \in \mathbb{R}^{R_{n-1}\times I_n \times R_n}, \forall n \in [1,N]$, that is
\begin{equation}\label{TR_model}
	\boldsymbol{\mathcal{L}}(i_1, i_2, \cdots, i_N) = \operatorname{Tr} \left(\mathbf{Z}_{1}(i_1)\mathbf{Z}_{2}(i_2)\cdots\mathbf{Z}_{N}(i_N)\right),  
\end{equation}
where $\operatorname{Tr}(\cdot)$ denotes the trace operation, $\mathbf{Z}_{n}(i_n) \in \mathbb{R}^{R_{n-1}\times R_n}$ is the $i_n$-th lateral slice of $\boldsymbol{\mathcal{Z}}^{(n)}$. $(R_0,R_1,\cdots,R_N)$ denotes the TR rank with $R_0 = R_N$. For simplicity, the TR decomposition of tensor $\boldsymbol{\mathcal{L}}$ is abbreviated by $\boldsymbol{\mathcal{L}} = \mathcal{R}(\boldsymbol{\mathcal{Z}}^{(1)},\boldsymbol{\mathcal{Z}}^{(2)}\cdots, \boldsymbol{\mathcal{Z}}^{(N)})$.
\begin{theorem}[Circular Dimensional Permutation Invariance \cite{zhao2016tensor}]
	\emph{For a TR format tensor $\boldsymbol{\mathcal{L}} \in \mathbb{R}^{I_1 \times I_2 \times \cdots \times I_N}$, the circular dimensional permutation $\overleftarrow{\boldsymbol{\mathcal{L}}^{n}} \in \mathbb{R}^{I_{n+1} \times \cdots \times I_{N} \times I_{1} \times \cdots \times I_{n}}$is defined by}
\end{theorem}
\begin{equation}
	\begin{split}
		&\overleftarrow{\boldsymbol{\mathcal{L}}^{n}}\left(i_{n+1},\cdots, i_{N}, i_{1}, \cdots, i_{n}\right)\\
		&=\operatorname{Tr}\left(\mathbf{Z}_{n+1}\left(i_{n+1}\right) \cdots \mathbf{Z}_{N}\left(i_{N}\right) \mathbf{Z}_{1}\left(i_{1}\right) \cdots \mathbf{Z}_{n}\left(i_{n}\right)\right).
	\end{split}
\end{equation}
Therefore, the circular dimensional permutation $\overleftarrow{\boldsymbol{\mathcal{L}}^{n}} = \mathcal{R}(\boldsymbol{\mathcal{Z}}^{(n+1)},\cdots,\boldsymbol{\mathcal{Z}}^{(N)},\boldsymbol{\mathcal{Z}}^{(1)},\cdots, \boldsymbol{\mathcal{Z}}^{(n)})$.
\begin{definition}[Tensor Connection Product(TCP) \cite{cichocki2014era}]
	\emph{Given a sequence of tensors $\boldsymbol{\mathcal{Z}}^{(n)} \in \mathbb{R}^{R_{n-1}\times I_{n} \times R_{n}}, \forall n \in [1,N]$, the tensor connection product is defined by}
\end{definition}
\begin{equation}
	\begin{split}
		\boldsymbol{\mathcal{Z}} = \operatorname{TCP}\left(\boldsymbol{\mathcal{Z}}^{(1)}, 	\boldsymbol{\mathcal{Z}}^{(2)},\cdots, \boldsymbol{\mathcal{Z}}^{(N)}\right) \in \mathbb{R}^{R_{0} \times (I_{1}I_{2}\cdots I_{N}) \times R_{N}}.
	\end{split}
\end{equation}
Specifically, the tensor connection product over all latent tensors expect $n$-th tensor is denoted as
\begin{equation}\label{expect_n}
	\begin{split}
		\boldsymbol{\mathcal{Z}}^{(\backslash n)} = & \operatorname{TCP}\left(\boldsymbol{\mathcal{Z}}^{(n+1)}, \cdots, \boldsymbol{\mathcal{Z}}^{(N)},\boldsymbol{\mathcal{Z}}^{(1)}, \cdots, \boldsymbol{\mathcal{Z}}^{(n-1)}\right) \\
		&\in \mathbb{R}^{R_{n} \times (I_{n+1}\cdots I_{N}I_{1}\cdots I_{n-1}) \times R_{n-1}}.
	\end{split}
\end{equation}
\section{Bayesian Robust Tensor Ring Decomposition}
\label{BRTRD}
In this section, we introduce our Bayesian robust tensor ring model and develop a variational Bayesian inference algorithm. 
\subsection{Model Specification}
Given an incomplete tensor $\boldsymbol{\mathcal{Y}}_{\Omega} \in \mathbb{R}^{I_1 \times I_2 \times \cdots \times I_N}$, its observed entries can be defined by $\{\boldsymbol{\mathcal{Y}}_{i_1 i_2 \cdots i_N}\mid(i_1, i_2, \cdots, i_N) \in \Omega\}$, where $\Omega$ is a set of indices. In addition, we define an indicator tensor $\boldsymbol{\mathcal{O}} \in \mathbb{R}^{I_1 \times I_2 \times \cdots \times I_N}$ whose entry $\boldsymbol{\mathcal{O}}_{i_1 i_2 \cdots i_N}$ is the same as one if $(i_1, i_2, \cdots, i_N) \in \Omega$ and zero otherwise. In the noisy measurement, we assume that $\boldsymbol{\mathcal{Y}}$ can be divided into three parts:
\begin{equation}
	\begin{split}
		\boldsymbol{\mathcal{Y}} = \boldsymbol{\mathcal{L}} + \boldsymbol{\mathcal{S}} + \boldsymbol{\mathcal{M}}, \text { s.t., } \boldsymbol{\mathcal{P}}_{\Omega}(\boldsymbol{\mathcal{L}}+\boldsymbol{\mathcal{S}} + \boldsymbol{\mathcal{M}})=\boldsymbol{\mathcal{P}}_{\Omega}(\boldsymbol{\mathcal{Y}}),
	\end{split}
\end{equation}
where $\boldsymbol{\mathcal{L}}$ denotes a low-rank component that can be generated by Eq. (\ref{TR_model}), $\boldsymbol{\mathcal{S}}$ is a sparse tensor that denotes gross noise, and $\boldsymbol{\mathcal{M}}$ is isotropic Gaussian noise. $\boldsymbol{\mathcal{P}}_{\Omega}$ is a linear projection such that the entries in the set $\Omega$ are given while the remaining entries are missing. Since $\boldsymbol{\mathcal{L}}$ is assumed to be a TR structure, a generative model from observed data under a probabilistic framework is defined by

\begin{footnotesize}
	\begin{equation}
		\begin{split}
			&p\left(\boldsymbol{\mathcal{Y}}_{\Omega} \mid\left\{\boldsymbol{\mathcal{Z}}^{(n)}\right\}_{n=1}^{N}, \boldsymbol{\mathcal{S}}_{\Omega}, \tau\right) \\
			& \!=\! \prod_{i_{1}}^{I_{1}} \!\cdots\! \prod_{i_{N}}^{I_{N}} \! \mathcal{N} \!\left(\boldsymbol{\mathcal{Y}}_{i_{1} \cdots i_{N}} \!\mid\! \mathcal{R}(\mathbf{Z}_{1}(i_{1}), \cdots, \mathbf{Z}_{N}(i_{N}))\!+\!\boldsymbol{\mathcal{S}}_{i_{1}  \cdots i_{N}}, \tau^{-1}\right)^{O_{i_{1} \cdots i_{N}}},
		\end{split}	
	\end{equation}
\end{footnotesize}	 
where $\tau$ denotes the noise precision. Since each core tensor $\boldsymbol{\mathcal{Z}}^{(n)}$ shares dimensionality with the other two core tensors, it requires two hyperparameters $\mathbf{u}_{r_{n-1}}^{(n-1)}, \mathbf{u}_{r_{n}}^{(n)}$ to involve in the component in $\boldsymbol{\mathcal{Z}}^{(n)}$. Hence, for core tensors $\{\boldsymbol{\mathcal{Z}}^{(n)}\}_{n=1}^{N}$, we place Gaussian distributions with zero mean and noise precision $\mathbf{u}_{r_{n-1}}^{(n-1)} * \mathbf{u}_{r_{n}}^{(n)}$ over them,
\begin{equation}\label{core_proba}
	\begin{split}
		&p\left(\boldsymbol{\mathcal{Z}}^{(n)} \mid \mathbf{u}^{(n-1)}, \mathbf{u}^{(n)}\right)\\
		&\!=\!\prod_{i_{n}}^{I_{n}} \prod_{r_{n-1}}^{R_{n-1}} \prod_{r_{n}}^{R_{n}}\!\mathcal{N}\!\left(\boldsymbol{\mathcal{Z}}^{(n)}\left(r_{n-1}, i_{n}, r_{n}\right)\! \mid\! 0,\left(\mathbf{u}_{r_{n-1}}^{(n-1)} * \mathbf{u}_{r_{n}}^{(n)}\right)^{-1}\right),\\
		& \forall n \in [1,N], 
	\end{split}
\end{equation}
where $\mathbf{u}^{(n)} = [u_{1}^{(n)}, \cdots, u_{R_n}^{(n)}]$. Further, we can rewrite Eq. (\ref{core_proba}) in a concise form as
\begin{equation}
	\begin{split}
		&p\left(\boldsymbol{\mathcal{Z}}^{(n)} \mid \mathbf{u}^{(n-1)}, \mathbf{u}^{(n)}\right) \\
		&= \prod_{i_{n}}^{I_{n}}\mathcal{MN}\left(\operatorname{vec}(\mathbf{Z}_{n}(i_n)) \mid 0,\left(\mathbf{U}^{(n)} \otimes \mathbf{U}^{(n-1)}\right)^{-1}\right),\\
		& \forall n \in [1,N], 
	\end{split}
\end{equation}
where $\mathbf{U}^{(n)} = \operatorname{diag}(\mathbf{u}^{(n)})$ denotes the precision matrix. In addition, we place Gamma distributions over $\{\mathbf{u}^{(n)}\}_{n=1}^{N}$, that is 
\begin{equation}
	\begin{aligned}
		&p\left(\mathbf{u}^{(n)} \mid \mathbf{c}^{(n)}, \mathbf{d}^{(n)}\right)=\prod_{r_{n}=1}^{R_{n}} \mathrm{Ga}\left(u_{r_{n}}^{(n)} \mid c_{r_{n}}^{(n)}, d_{r_{n}}^{(n)}\right), \\
		&\forall n \in [1,N],
	\end{aligned}
\end{equation}
where parameters $\mathbf{c}^{(n)} = [c_{1}^{(n)}, \cdots, c_{R_n}^{(n)}]$ and $\mathbf{d}^{(n)} = [d_{1}^{(n)}, \cdots, d_{R_n}^{(n)}]$.
\begin{figure}[t]  
	\centering  
	\includegraphics[width=1\linewidth]{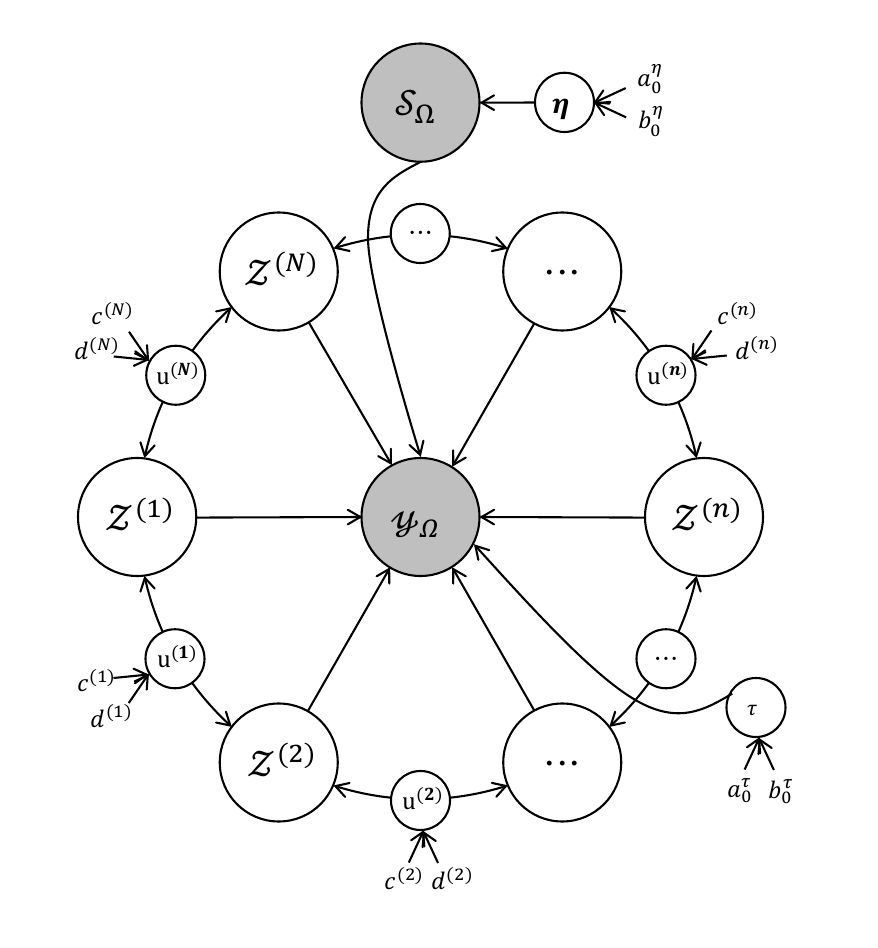}
	\caption{The graphical illustration of BRTR model.}
	\label{fig_BRTR} 
\end{figure}

Each entry of the sparse tensor $\boldsymbol{\mathcal{S}}$ is modeled as a Gaussian distribution with zero mean and noise precision $\boldsymbol{\eta}$, which can be represented by
\begin{equation}
	p\left(\boldsymbol{\mathcal{S}}_{\Omega} \mid \boldsymbol{\eta}\right)=\prod_{i_{1} \cdots i_{N}} \mathcal{N}\left(\boldsymbol{\mathcal{S}}_{i_{1} \cdots i_{N}} \mid 0, \eta_{i_{1} \cdots i_{N}}^{-1}\right)^{\boldsymbol{\mathcal{O}}_{i_{1} \cdots i_{N}}}. 
\end{equation} 
For the noise precision $\boldsymbol{\eta}$, we place a Gamma distribution over it, that is
\begin{equation}
	p(\boldsymbol{\eta})=\prod_{i_{1} \cdots i_{N}} \mathrm{Ga}\left(\eta_{i_{1}\cdots i_{N}} \mid a_{0}^{\eta}, b_{0}^{\eta}\right).
\end{equation}

To complete our model, we place a Gamma distribution over the noise precision $\tau$, that is
\begin{equation}
	p(\tau) = \mathrm{Ga}\left(\tau \mid a_{0}^{\tau}, b_{0}^{\tau}\right).
\end{equation}

Finally, we can build a full probabilistic Bayesian robust tensor ring model, as shown in Fig. \ref{fig_BRTR}. The joint distribution of BRTR model is defined by
\begin{equation}\label{BRTR}
	\begin{split}
		&p(\boldsymbol{\mathcal{Y}}_{\Omega}, \boldsymbol{\mathcal{H}}) = 
		p\left(\boldsymbol{\mathcal{Y}}_{\Omega} \mid\left\{\boldsymbol{\mathcal{Z}}^{(n)}\right\}_{n=1}^{N}, \boldsymbol{\mathcal{S}}_{\Omega}, \tau\right)  \\
		&\times \prod_{n=1}^{N} p\left(\boldsymbol{\mathcal{Z}}^{(n)} \mid \mathbf{u}^{(n-1)}, \mathbf{u}^{(n)}\right) \prod_{n=1}^{N} p(\mathbf{u}^{(n)}) p\left(\boldsymbol{\mathcal{S}}_{\Omega} \mid \boldsymbol{\eta}\right)  p(\boldsymbol{\eta}) p(\tau), 
	\end{split}
\end{equation}
where $\boldsymbol{\mathcal{H}} = \{\{\boldsymbol{\mathcal{Z}}^{(n)}\}_{n=1}^{N}, \{\mathbf{u}\}_{n=1}^{N}, \boldsymbol{\mathcal{S}}_{\Omega}, \boldsymbol{\eta}, \tau\}$ denotes all unknown parameters. To infer the missing entries $\boldsymbol{\mathcal{Y}}_{\backslash \Omega}$ through known entries, the predictive distribution is expressed by
\begin{equation}
	p\left(\boldsymbol{\mathcal{Y}}_{\backslash \Omega} \mid \boldsymbol{\mathcal{Y}}_{\Omega}\right)=\int p\left(\boldsymbol{\mathcal{Y}}_{\Omega} \mid \boldsymbol{\mathcal{H}}\right)  p\left(\boldsymbol{\mathcal{H}} \mid  \boldsymbol{\mathcal{Y}}_{\Omega}\right) \mathrm{d} \boldsymbol{\mathcal{H}},
\end{equation}
where $p(\boldsymbol{\mathcal{H}} \mid  \boldsymbol{\mathcal{Y}}_{\Omega}) = p\left(\boldsymbol{\mathcal{H}}, \boldsymbol{\mathcal{Y}}_{\Omega}\right)/ \int p\left(\boldsymbol{\mathcal{H}}, \boldsymbol{\mathcal{Y}}_{\Omega}\right)\mathrm{d} \boldsymbol{\mathcal{H}}$.
\subsection{Variational Bayesian Inference for Model Learning}
For the BRTR model, it is difficult to obtain an exact solution. Fortunately, the variational Bayesian inference algorithm \cite{tipping2001sparse} provides a closed-form posterior approximation and is computationally efficient for the above problem. VB inference aims to find a distribution $q(\boldsymbol{\mathcal{H}})$ that approximates the true distribution $p(\boldsymbol{\mathcal{H}} \mid \boldsymbol{\mathcal{Y}}_{\Omega})$ by minimizing the Kullback-Leibler (KL) divergence between $q(\boldsymbol{\mathcal{H}})$ and $p(\boldsymbol{\mathcal{H}} \mid \boldsymbol{\mathcal{Y}}_{\Omega})$, which can be expressed by
\begin{equation}
	\begin{split}
		\mathrm{KL}(q(\boldsymbol{\mathcal{H}}) &\| p(\boldsymbol{\mathcal{H}} \mid \boldsymbol{\mathcal{Y}}_{\Omega}))  = \int q(\boldsymbol{\mathcal{H}}) \ln \left\{\frac{q(\boldsymbol{\mathcal{H}})}{p(\boldsymbol{\mathcal{H}} \mid \boldsymbol{\mathcal{Y}}_{\Omega})}\right\} \mathrm{d} \boldsymbol{\mathcal{H}}\\
		& = \ln p(\boldsymbol{\mathcal{Y}}_{\Omega})-\int q(\boldsymbol{\mathcal{H}}) \ln 	\left\{\frac{p(\boldsymbol{\mathcal{Y}}_{\Omega},\boldsymbol{\mathcal{H}})}{q(\boldsymbol{\mathcal{H}})}\right\} \mathrm{d} \boldsymbol{\mathcal{H}}.
	\end{split}
\end{equation}
The problem of minimizing KL divergence is equivalent to maximizing  $\boldsymbol{\mathscr{L}(q)} = \int q(\boldsymbol{\mathcal{H}}) \ln 	\left\{{p(\boldsymbol{\mathcal{Y}}_{\Omega},\boldsymbol{\mathcal{H}})}/{q(\boldsymbol{\mathcal{H}})}\right\} \mathrm{d} \boldsymbol{\mathcal{H}}$ since $\ln p(\boldsymbol{\mathcal{Y}}_{\Omega}) $ is a constant. According to mean-field approximation, the approximate distribution $q(\boldsymbol{\mathcal{H}})$ can be expressed by
\begin{equation}
	q(\boldsymbol{\mathcal{H}})=\prod_{n=1}^{N} q\left(\boldsymbol{\mathcal{Z}}^{(n)}\right)\prod_{n=1}^{N} q\left(\mathbf{u}^{(n)}\right) q\left(\boldsymbol{\mathcal{S}}_{\Omega}\right) q(\boldsymbol{\eta}) q(\tau).
\end{equation}
Thus, we can update the $j$-th factor by
\begin{equation}
	\ln q_{j}\left(\boldsymbol{\mathcal{H}}_{j}\right)=\mathbb{E}_{q_{\left(\boldsymbol{\mathcal{H}} \backslash \boldsymbol{\mathcal{H}}_{j}\right)}}\left[\ln p\left(\boldsymbol{\mathcal{Y}}_{\Omega}, \boldsymbol{\mathcal{H}}\right)\right]+\text { const},
\end{equation}
where $\mathbb{E}_{q_{\left(\boldsymbol{\mathcal{H}} \backslash \boldsymbol{\mathcal{H}}_{j}\right)}}[\cdot]$ denotes an expectation with respect to $q$ distribution over all variables in $\boldsymbol{\mathcal{H}}$ expect $\boldsymbol{\mathcal{H}}_{j}$. For convenience, we use $\mathbb{E}[\cdot]$ to denote the expectation with respect to $q(\boldsymbol{\mathcal{H}})$.
\subsubsection{Posterior Distribution of Core Tensors $\{\boldsymbol{\mathcal{Z}}^{(n)}\}_{n=1}^{N}$}
The posterior distribution of core tensors $\{\boldsymbol{\mathcal{Z}}^{(n)}\}_{n=1}^{N}$ are assumed to obey Gaussian distributions, which can be expressed by
\begin{equation}
	q\left(\boldsymbol{\mathcal{Z}}^{(n)}\right)=\prod_{i_{n}=1}^{I_{n}} \mathcal{MN}\left(\operatorname{vec}\left(\mathbf{Z}_{n}\left(i_{n}\right)\right) \mid \tilde{\mathbf{z}}^{(n)}_{i_{n}}, \mathbf{V}_{i_{n}}^{n}\right),
\end{equation}
where mean and variance can be updated by
\begin{small}
	\begin{equation}\label{update_Z}
		\begin{aligned}
			&\tilde{\mathbf{z}}^{(n)}_{i_{n}}=\mathbb{E}[\tau] \mathbf{V}_{i_{n}}^{(n)} \mathbb{E}\left[\left(\boldsymbol{\mathcal{Z}}^{(\backslash n)}_{\mathbb{O}_{i_n}}\right)^{\top}\right] \left(\boldsymbol{\mathcal{Y}}_{\mathbb{O}_{i_n}} - \mathbb{E}[\boldsymbol{\mathcal{S}}_{\mathbb{O}_{i_n}}]\right),\\
			&\mathbf{V}_{i_{n}}^{n}=\left(\mathbb{E}[\tau] \mathbb{E}\left[\left(\boldsymbol{\mathcal{Z}}^{(\backslash n)}_{\mathbb{O}_{i_n}}\right)^{\top} \boldsymbol{\mathcal{Z}}^{(\backslash n)}_{\mathbb{O}_{i_n}}\right]+\mathbb{E}\left[\left(\mathbf{U}^{(n)} \otimes \mathbf{U}^{(n-1)}\right)\right]\right)^{-1},	
		\end{aligned}
	\end{equation}
\end{small}
\renewcommand{\algorithmicrequire}{\textbf{Input:}} 
\renewcommand{\algorithmicensure}{\textbf{Initialization:}}
\begin{algorithm}[t] 
	\caption{Bayesian Robust Tensor Ring Decomposition} 
	\label{alg1} 
	\begin{algorithmic} 
		\REQUIRE An observed tensor $\boldsymbol{\mathcal{Y}}_{\Omega} $, an indicator tensor $\boldsymbol{\mathcal{O}}$ and max TR rank $(R_0,R_1,\cdots R_N)$.\\
		\ENSURE $\boldsymbol{\mathcal{Z}}^{(n)}, \mathbf{V}^{(n)}, \mathbf{u}^{(n)}, \forall n \in [1,N], \tilde{\boldsymbol{\mathcal{S}}}, \sigma^2, \boldsymbol{\eta}, \tau, $ top level hyperparameters $c^{(n)}$, $d^{(n)}, \forall n \in [1,N]$, $a^{\eta}, b^{\eta}, a^{\tau}$ and $b^{\tau}$. \\
		\REPEAT 
		\FOR{$n=1$ to $N$}
		\STATE Update $q(\boldsymbol{\mathcal{Z}}^{(n)})$ using Eq. (\ref{update_Z}); 
		\ENDFOR
		\FOR{$n=1$ to $N$}
		\STATE Update $q(\mathbf{u}^{(n)})$ using Eq. (\ref{update_u});
		\ENDFOR
		\STATE Update $q(\boldsymbol{\mathcal{S}})$ using Eq. (\ref{update_S});
		\STATE Update $q(\boldsymbol{\eta})$ using Eq. (\ref{update_eta});
		\STATE Update $q(\tau)$ using Eq. (\ref{update_tau});
		\STATE Model reduction by eliminating zero-components in
		$\{\boldsymbol{\mathcal{Z}}^{(n)}\}_{n=1}^{N}$; 
		\UNTIL{convergence} 
	\end{algorithmic} 
\end{algorithm}

where $\boldsymbol{\mathcal{Y}}_{\mathbb{O}_{i_n}} \in \mathbb{R}^{{(I_{n+1}\cdots I_{N}I_{1}\cdots I_{n-1})}_{\mathbb{O}_{i_n}}\times 1} $ denotes partial observed data and $\boldsymbol{\mathcal{S}}_{\mathbb{O}_{i_{n}}} \in \mathbb{R}^{{(I_{n+1}\cdots I_{N}I_{1}\cdots I_{n-1})}_{\mathbb{O}_{i_n}}\times 1}$ denotes the noise of partial observed data $\boldsymbol{\mathcal{Z}}^{(\backslash n)}_{\mathbb{O}_{i_n}} \in \mathbb{R}^{{(I_{n+1}\cdots I_{N}I_{1}\cdots I_{n-1})}_{\mathbb{O}_{i_n}}\times R_{n-1}R_{n}}$ is defined by Eq. (\ref{expect_n}), $\mathbb{O}_{i_n}$ denotes the observed entries in $\boldsymbol{\mathcal{Y}}_{\mathbb{O}_{i_n}}$.
\subsubsection{Posterior Distribution of Hyperparameters $\{\mathbf{u}^{(n)}\}_{n=1}^{N}$}
The posterior distribution of hyperparameters $\{\mathbf{u}^{(n)}\}_{n=1}^{N}$ are assumed to obey Gamma distribution, which can be expressed by
\begin{equation}
	q\left(\mathbf{u}^{(n)}\right)=\prod_{r_{n}=1}^{R_{n}} \mathrm{Ga}\left(\mathbf{u}_{r_{n}}^{(n)} \mid \tilde{c}^{(n)}_{r_{n}}, \tilde{d}^{(n)}_{r_{n}}\right),
\end{equation}
where parameters $\tilde{c}^{(n)}_{r_{n}}$ and $\tilde{d}^{(n)}_{r_{n}}$ can be updated by
\begin{equation}\label{update_u}
	\begin{split}
		\tilde{c}^{(n)}_{r_{n}} &=  c^{(n)}_{r_{n}} +\frac{1}{2}\left(I_{n} R_{n-1}+I_{n+1} R_{n+1}\right), \\
		\tilde{d}^{(n)}_{r_{n}} &=  d^{(n)}_{r_{n}}+\frac{1}{2}\operatorname{Tr}\left(\mathbb{E}\left[\mathbf{U}^{(n-1)}\right] \mathbb{E}\left[\mathbf{Z}_{n}\left(r_{n}\right) \left(\mathbf{Z}_{n}\left(r_{n}\right)\right)^{\top}\right]\right)\\
		&+\frac{1}{2}\operatorname{Tr}\left(\mathbb{E}\left[\mathbf{U}^{(n+1)}\right] \mathbb{E}\left[\left(\mathbf{Z}_{n+1}\left(r_{n}\right)\right)^{\top} \mathbf{Z}_{n+1}\left(r_{n}\right)\right]\right),
	\end{split}
\end{equation}
where $\mathbf{Z}_{n} (r_n) = \boldsymbol{\mathcal{Z}}^{(n)}(:,:,r_n) \in \mathbb{R}^{R_{n-1} \times I_{n}}$ and $\mathbf{Z}_{n+1} (r_n) = \boldsymbol{\mathcal{Z}}^{(n+1)}(r_{n+1},:,:) \in \mathbb{R}^{I_{n+1} \times R_{n+1}}$. 
\subsubsection{Posterior Distribution of Sparse Tensor $\boldsymbol{\mathcal{S}}$}
The posterior distribution of sparse tensor is assumed to obey a Gaussian distribution, that is
\begin{equation}
	q(\boldsymbol{\mathcal{S}})=\prod_{\left(i_{1}, \cdots, i_{N}\right) \in \Omega} \mathcal{N}\left(\boldsymbol{\mathcal{S}}_{i_{1}  \cdots i_{N}} \mid \tilde{\boldsymbol{\mathcal{S}}}_{i_{1} \cdots i_{N}}, \sigma_{i_{1} \cdots i_{N}}^{2}\right)
\end{equation}
where mean and variance can be updated by
\begin{equation}\label{update_S}
	\begin{aligned}
		\tilde{\boldsymbol{\mathcal{S}}}_{i_{1} \cdots i_{N}} &=\sigma_{i_{1} \cdots i_{N}}^{2} \mathbb{E}[\tau]\left(\boldsymbol{\mathcal{Y}}_{i_{1}  \cdots i_{N}}-\mathbb{E}\left[\mathcal{R} \left(\mathbf{Z}_{1}(i_1)\cdots\mathbf{Z}_{N}(i_N)\right)\right]\right), \\
		\sigma_{i_{1} \cdots i_{N}}^{2} &=\left(\mathbb{E}\left[\eta_{i_{1}\cdots i_{N}}\right]+\mathbb{E}[\tau]\right)^{-1}.
	\end{aligned}
\end{equation}

\begin{table*}[t]
	\centering
	\caption{The performance of both BRTR and RTRC methods under different MR and SR in a noise-free environment.}
	\begin{tabular}{c|c|c|c|c|c|c|c|c}
		\specialrule{.08em}{0pt}{0pt}
		\multicolumn{4}{c|}{Method}                                                                         & \multicolumn{2}{c|}{RTRC} & \multicolumn{3}{c}{BRTR}    \\ \specialrule{.08em}{0pt}{0pt}
		I                                  & R                              & MR                    & SR   & $\frac{\left\|\hat{\boldsymbol{\mathcal{L}}}-\boldsymbol{\mathcal{L}}\right\|_{F}}{\left\|\boldsymbol{\mathcal{L}}\right\|_{F}}$      & $\frac{\left\|\hat{\boldsymbol{\mathcal{S}}}-\boldsymbol{\mathcal{S}}\right\|_{F}}{\left\|\boldsymbol{\mathcal{S}}\right\|_{F}}$     & REE & $\frac{\left\|\hat{\boldsymbol{\mathcal{L}}}-\boldsymbol{\mathcal{L}}\right\|_{F}}{\left\|\boldsymbol{\mathcal{L}}\right\|_{F}}$   & $\frac{\left\|\hat{\boldsymbol{\mathcal{S}}}-\boldsymbol{\mathcal{S}}\right\|_{F}}{\left\|\boldsymbol{\mathcal{S}}\right\|_{F}}$  \\ \specialrule{.08em}{0pt}{0pt}
		\multirow{6}{*}{{[}10,10,10,10{]}} & \multirow{6}{*}{{[}3,3,3,3{]}} & \multirow{2}{*}{0\%}  & 10\% & 1.51e-05    & 1.18e-05   & 0.000  & \textbf{3.84e-08} & \textbf{2.79e-08} \\ \cline{4-9}
		&                                &                       & 15\% & 3.30e-01    & 2.30e-01   & 0.000  & \textbf{2.04e-07} & \textbf{1.47e-07} \\ \cline{3-9}
		&                                & \multirow{2}{*}{10\%} & 10\% & 9.11e-07    & 3.11e-01   & 0.050  & \textbf{2.70e-08} & \textbf{3.07e-01} \\ \cline{4-9}
		&                                &                       & 15\% & 5.23e-01    & 4.42e-01   & 0.050  & \textbf{2.09e-07} & \textbf{3.16e-01} \\ \cline{3-9}
		&                                & \multirow{2}{*}{20\%} & 10\% & 6.72e-02    & 4.51e-01   & 0.075  & \textbf{8.07e-08} & \textbf{4.48e-01} \\ \cline{4-9}
		&                                &                       & 15\% & 5.56e-01    & 5.59e-01   & 0.075  & \textbf{9.47e-08} & \textbf{4.43e-01} \\ \specialrule{.08em}{0pt}{0pt}
		\multirow{6}{*}{{[}10,10,10,10{]}} & \multirow{6}{*}{{[}3,2,3,2{]}} & \multirow{2}{*}{0\%}  & 10\% & 2.66e-07    & 1.67e-07   & 0.250  & \textbf{2.41e-07} & \textbf{1.64e-07} \\ \cline{4-9}
		&                                &                       & 15\% & 3.34e-01    & 1.92e-01   & 0.150  & \textbf{1.17e-07} & \textbf{7.50e-08} \\ \cline{3-9}
		&                                & \multirow{2}{*}{10\%} & 10\% & 7.50e-07    & \textbf{3.10e-01}   & 0.250  & \textbf{1.22e-07} & 3.39e-01 \\ \cline{4-9}
		&                                &                       & 15\% & 3.95e-01    & 9.40e-01   & 0.100  & \textbf{1.71e-07} & \textbf{3.19e-01} \\ \cline{3-9}
		&                                & \multirow{2}{*}{20\%} & 10\% & 1.77e-06    & 4.59e-01   & 0.250  & \textbf{1.81e-07} & \textbf{4.45e-01} \\ \cline{4-9}
		&                                &                       & 15\% & 4.86e-01    & 5.21e-01   & 0.250  & \textbf{9.42e-08} & \textbf{4.45e-01} \\ \specialrule{.08em}{0pt}{0pt}
	\end{tabular}
	\label{tab_syn2}
\end{table*}

\subsubsection{Posterior Distribution of Hyperparameter $\boldsymbol{\eta}$}
The posterior distribution of hyperparameter $\boldsymbol{\eta}$ is assumed to obey a Gamma distribution of each entry, which can be expressed by
\begin{equation}
	q(\boldsymbol{\eta})=\prod_{\left(i_{1}, \cdots, i_{N}\right) \in \Omega} \mathrm{Ga}\left(\eta_{i_{1} \cdots i_{N}} \mid a_{_{i_{1} \cdots i_{N}}}^{\eta}, b_{_{i_{1} \cdots i_{N}}}^{\eta}\right),
\end{equation}
where parameters $a^{\eta}$ and $ b^{\eta}$ can be updated by
\begin{equation}\label{update_eta}
	a_{{i_{1} \cdots i_{N}}}^{\eta}=a_{0}^{\eta}+\frac{1}{2}, \quad b_{{i_{1} \cdots i_{N}}}^{\eta}=b_{0}^{\eta}+\frac{1}{2}\left(\tilde{\boldsymbol{\mathcal{S}}}_{i_{1} \cdots i_{N}}^{2}+\sigma_{i_{1} \cdots i_{N}}^{2}\right).
\end{equation}
\subsubsection{Posterior Distribution of Hyperparameter $\tau$}
The posterior distribution of hyperparameter $\tau$ is assumed to obey a Gamma distribution of each entry, that is  
\begin{equation}
	q(\tau)=\operatorname{Ga}\left(\tau \mid a^{\tau}, b^{\tau}\right),
\end{equation}
where parameters $a^{\tau}$ and $b^{\tau}$ can be updated by
\begin{equation}\label{update_tau}
	\begin{aligned}
		&a^{\tau}=a_{0}^{\tau}+\frac{1}{2} \sum_{i_{1} \cdots i_{N}} \boldsymbol{\mathcal{O}}_{i_{1} \cdots i_{N}}, \\
		&b^{\tau}=b_{0}^{\tau}+\frac{1}{2} \mathbb{E}\left[\left\|\boldsymbol{\mathcal{O}} \circledast\left(\boldsymbol{\mathcal{Y}}-\mathcal{R} \left(\mathbf{Z}_{1}(i_1)\cdots\mathbf{Z}_{N}(i_N)\right)-\boldsymbol{\mathcal{S}}\right)\right\|_{F}^{2}\right] .
	\end{aligned}
\end{equation}
Here we complete the derivation of the entire variational inference and the whole procedure of the BRTR method is summarized in Algorithm \ref{alg1}. The lower bound of model evidence in can be computed by
\begin{equation}
	\mathscr{L}(q)=\mathbb{E}_{q(\boldsymbol{\mathcal{H}})}[\ln p(\boldsymbol{\mathcal{Y}}_{\Omega}, \boldsymbol{\mathcal{H}})]+H(q(\boldsymbol{\mathcal{H}})),
\end{equation}
where $\mathbb{E}_{q(\boldsymbol{\mathcal{H}})}[\cdot]$ indicates the posterior expectation while $H(\cdot)$ represents the entropy. Since $\mathscr{L}(q)$ should not decrease in each iteration, we can use it to test convergence. A detailed derivation process can be found on the supplementary material. 

\subsection{Complexity Analysis}
Assuming that we have an $N$-order tensor $\boldsymbol{\mathcal{Y}} \in \mathbb{R}^{I_1 \times I_2 \cdots \times I_N}$ with TR rank $(R_0, R_1 \cdots, R_N)$ where $\{I_n\}_{n=1}^{N}$ = I and $\{R_n\}_{n=0}^{N}$ = R. The computation cost of updating core factor $\boldsymbol{\mathcal{Z}}^{(n)}$ is $O((N-1)|O_I| R^6|)$. The computation costs are $O(2IR^2)$ for hyperparameters $u^{(n)}$, $O(|O|NR)$ sparse tensor $\boldsymbol{\mathcal{S}}$, $O(|O|NR^6)$ for $\tau$ where $|O| = \sum_{(i_1,\cdots, i_N) \in \Omega}\boldsymbol{\mathcal{O}}_{i_1\cdots i_N}$ denotes the number of total observations. Therefore, the total complexity of our algorithm is $O(N(N-1)|O_I|R^6 + 2NIR^2 + |O|NR + |O|NR^6)$ in each iteration. 

\subsection{Initialization}
To achieve faster convergence, a good initialization is necessary. Therefore, we employ TR approximation \cite{wang2017efficient} to initial core tensors $\{\boldsymbol{\mathcal{Z}}^{(n)}\}_{n=1}^{N}$. For top level hyperparameters $\{c^{(n)}\}_{n=1}^{N}$, $\{d^{(n)}\}_{n=1}^{N}$, $a^{\eta}, b^{\eta}, a^{\tau}$ and $b^{\tau}$, we set them to $10^{-6}$ to induce a non-informative prior. In addition, the expectation of hyperparameters are set by $\mathbb{E}(\tau) = 10$, $\mathbb{E}(\eta_{i_1\cdots i_N}) = 1, \forall n \forall i_n$ and  $\mathbb{E}(\mathbf{u}^{(n)}) = \mathbf{I}$. The sparse tensor $\boldsymbol{\mathcal{S}}_{i_1\cdots i_N}$ is sampled form $\mathcal{N}(0,1)$ while $\sigma_{i_{1} \cdots i_{N}}^{2}$ equals $\mathbb{E}(\eta_{i_1\cdots i_N}^{-1})$. The maximum TR rank is set to $30*ones(1,N+1)$ in the following experiments. 
\section{Experiments}
\label{exper}
In this section, we conduct extensive experiments to evaluate the effectiveness of our BRTR method and compare it with several state-of-the-art (SOTA) methods, including BRTF \cite{zhao2015bayesian}, TRLRF \cite{yuan2019tensor}, TRNNM \cite{yu2019tensor}, SiLRTC-TT \cite{bengua2017efficient}, TNN \cite{wang2019robust, wang2020OITNN}, TTNN \cite{song2020robust} and RTRC \cite{huang2020robust}. All experimental settings are based on the default parameters suggested by the authors. All experiments are performed under different missing ratio (MR) and sparse noise ratio (SR). MR is defined by
\begin{equation}
	MR = \frac{M}{\prod_{n=1}^{N}I_n},
\end{equation}
where $M$ is the number of total missing entries. SR is defined by 
\begin{equation}
	SR = \frac{S}{|O|},
\end{equation}
where $S$ is the sparse noise ratio of observations. In addition, we conduct our experiments using MATLAB 2020b on a desktop computer with a 3.40GHz Inter(R) Core(TM) i7-6700 CPU and 16GB RAM. All experiments are repeated five times and averaged to eliminate random errors.

\subsection{Experiments on Synthetic Data}
\begin{figure}[t]
	\centering
	\subfigure{\includegraphics[width=1.6in]{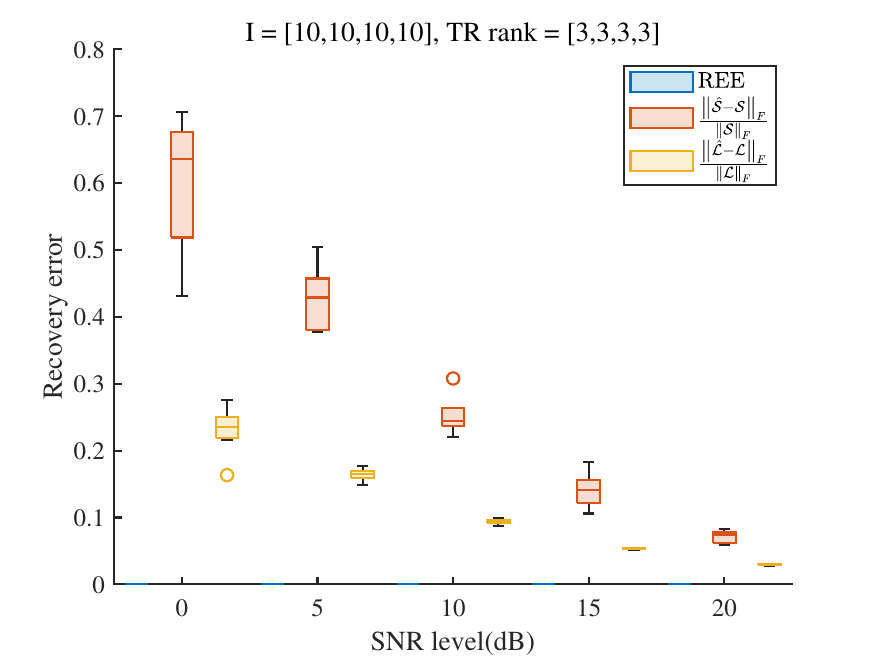}}
	\subfigure{\includegraphics[width=1.6in]{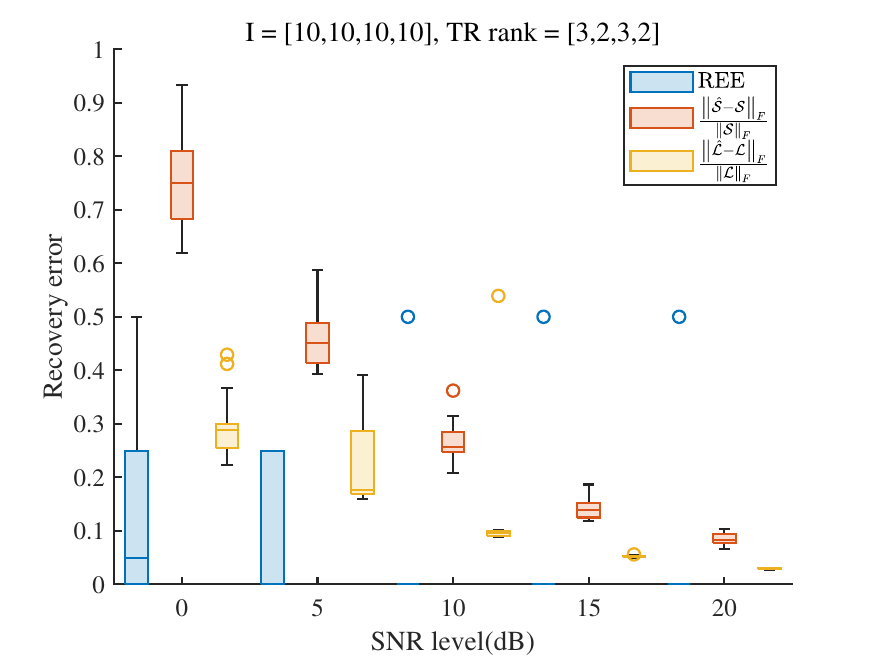}}
	\caption{The recovery performance and REE at different SNR level with MR = 0\% and SR = 10\%.}
	\label{fig_syn1}
\end{figure}

We first perform some experiments on the synthetic data to evaluate our BRTR method. We consider two experimental settings, i.e. $I = [10,10,10,10], R = [3,3,3,3]$ and $I = [10,10,10,10], R = [3,2,3,2]$ where $I$ is the size of dimension and $R$ is the TR rank with $R_0 = R_4$. The initial rank is set to the same dimension as the original data. The low-rank component $\boldsymbol{\mathcal{L}}$ is generated by $\boldsymbol{\mathcal{L}} = \mathcal{R}(\boldsymbol{\mathcal{Z}}^{(1)},\boldsymbol{\mathcal{Z}}^{(2)}, \boldsymbol{\mathcal{Z}}^{(3)},\boldsymbol{\mathcal{Z}}^{(4)})$ where each core tensor $\boldsymbol{\mathcal{Z}}^{(n)} \in \mathbb{E}^{R_{n-1} \times I_{n} \times R_{n}}, n \in [1,4]$ is sampled from a Gaussian distribution. The sparse component $\boldsymbol{\mathcal{S}}$ is randomly corrupted by outliers, which are drawn from a uniform distribution $\mathcal{U}(-|H|,|H|)$ where $H = \max(\operatorname{vec}(\mathcal{X}))$. The precision of noise component $\boldsymbol{\mathcal{M}}$ is controlled by the signal-to-noise ratio (SNR), that is 
\begin{equation}
	\operatorname{SNR} = 10 \log_{10} (\frac{\sigma_{\boldsymbol{\mathcal{L}}}}{\sigma_{\boldsymbol{\mathcal{M}}}}),
\end{equation}
where $\sigma_{\boldsymbol{\mathcal{L}}}$ and $\sigma_{\boldsymbol{\mathcal{M}}}$ denote the variance of $\boldsymbol{\mathcal{L}}$ and $\boldsymbol{\mathcal{M}}$, respectively. Thus, we can obtain the synthetic data $\boldsymbol{\mathcal{Y}} = \boldsymbol{\mathcal{O}} \circledast(\boldsymbol{\mathcal{L}} + \boldsymbol{\mathcal{S}} + \boldsymbol{\mathcal{M}})$. Our goal is to recover the low-rank component $\hat{\boldsymbol{\mathcal{L}}}$ and sparse component $\hat{\boldsymbol{\mathcal{S}}}$ and infer the correct TR rank $\hat{R}$ from the observed data $\boldsymbol{\mathcal{Y}}$ simultaneously. We also defined rank estimation error (REE), that is
\begin{equation}
	\operatorname{REE} = \frac{\sum_{n=1}^{N}|\hat{R}_{n} - R_{n}|}{N}.
\end{equation}
The first experiment is to test the recovery performance and REE at different SNR levels in a fully observed situation. Fig. \ref{fig_syn1} depicts the experiment results. From Fig. \ref{fig_syn1}, we can observe that: (1) The recovery error ${\|\hat{\boldsymbol{\mathcal{L}}}-\boldsymbol{\mathcal{L}}\|_{F}}/{\|\boldsymbol{\mathcal{L}}\|_{F}}$,  ${\|\hat{\boldsymbol{\mathcal{S}}}-\boldsymbol{\mathcal{S}}\|_{F}}/{\|\boldsymbol{\mathcal{S}}\|_{F}}$, and REE decrease as SNR level increases. (2) BRTR infers the correct TR rank more accurately in the case of balanced rank. (3) When the TR rank is unbalanced, BRTR infers the TR rank with a small error and the recovery performance is unsatisfactory at lower SNR. However, at higher SNR, BRTR is still able to infer the correct TR rank and has a fine recovery performance. Finally, we can see that BRTR always learns the correct TR rank and has a small recovery error in most cases at higher SNR. 
\begin{figure}[t]  
	\centering  
	\includegraphics[width=1\linewidth]{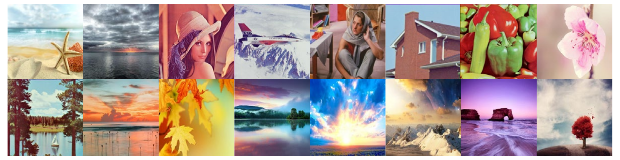}
	\caption{Testing color images.}
	\label{fig_color_image} 
\end{figure}
To further verify the effectiveness of the proposed BRTR method, we compare it with the RTRC method, which is a robust low-rank TR completion model in a noise-free environment. Therefore, for the sake of fairness, we consider evaluating the performance of both BRTR and RTRC methods under different MR and SR in a noise-free environment, as shown in TABLE \ref{tab_syn2}. From TABLE \ref{tab_syn2}, we can obtain that both RTRC and BRTR methods have a better recovery performance when MR = 0\%. As MR increases, RTRC has a performance degradation, especially at higher SR. For BRTR, it can always accurately recover low-rank component but has a larger recovery error when reconstructing sparse component. In addition, BRTR is more difficult to infer the correct TR rank as MR increases. In the case of a balanced rank, BRTR has a smaller REE compared to the unbalanced rank. In general, BRTR has a superior recovery performance compared to RTRC. It is worth noting that the TR rank of BRTR is automatically determined from the synthetic data whereas the TR rank of RTRC must be set manually.

\subsection{Experiments on Color Images}
In this section, we perform image recovery and denoising experiments to evaluate our proposed BRTR method. We choose some popular images with the size $256 \times 256 \times 3$, as shown in Fig. \ref{fig_color_image}. In each image, 30\% pixels are observed and 10\% pixels entries from the observation are set to random values in [0,255]. 

\begin{figure}[t]  
	\centering  
	\includegraphics[width=1\linewidth]{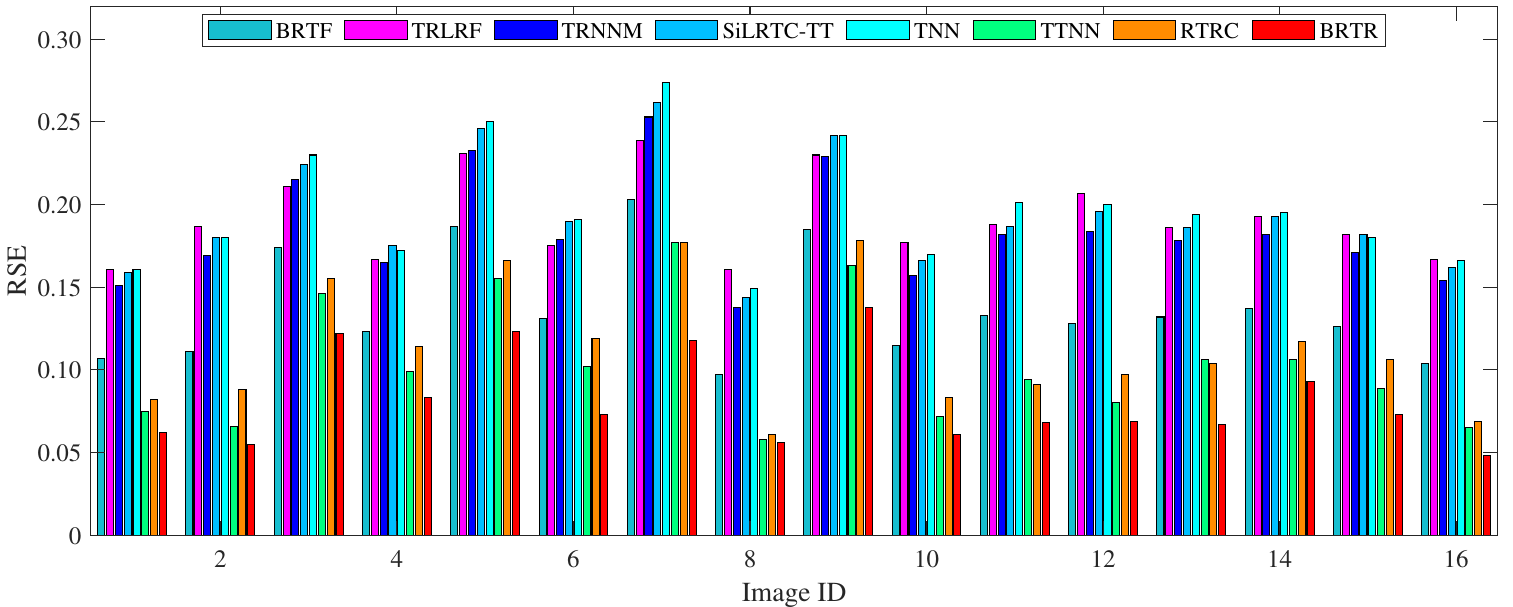}
	\caption{The RSE on testing color images with MR = 70\% and SR = 10\% over BRTF, TRLRF, TRNNM, SiLRTC-TT, TNN, TTNN, RTRC and BRTR methods.}
	\label{fig_color_rse} 
\end{figure}
\begin{figure}[t]  
	\centering  
	\includegraphics[width=1\linewidth]{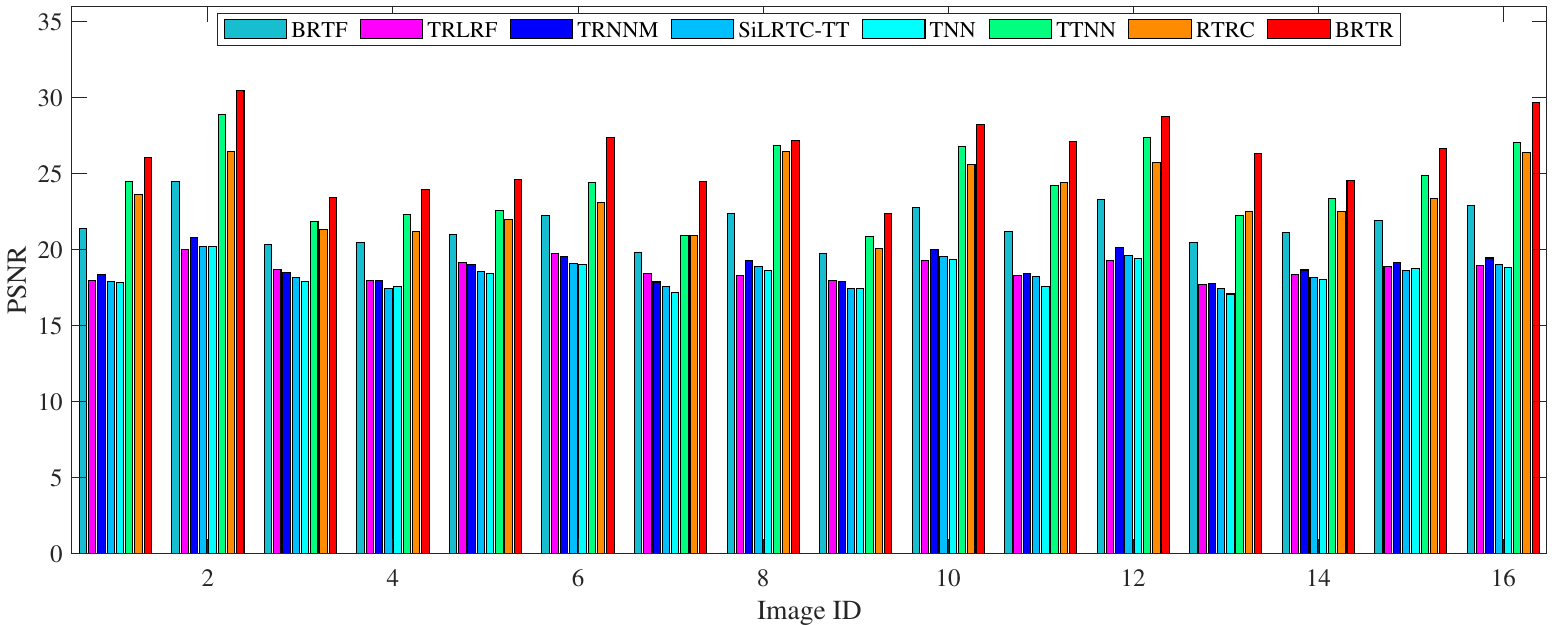}
	\caption{The PSNR on testing color images with MR = 70\% and SR = 10\% over BRTF, TRLRF, TRNNM, SiLRTC-TT, TNN, TTNN, RTRC and BRTR methods.}
	\label{fig_color_psnr} 
\end{figure}
\begin{figure*}[h]  
	\centering  
	\includegraphics[width=0.8\linewidth]{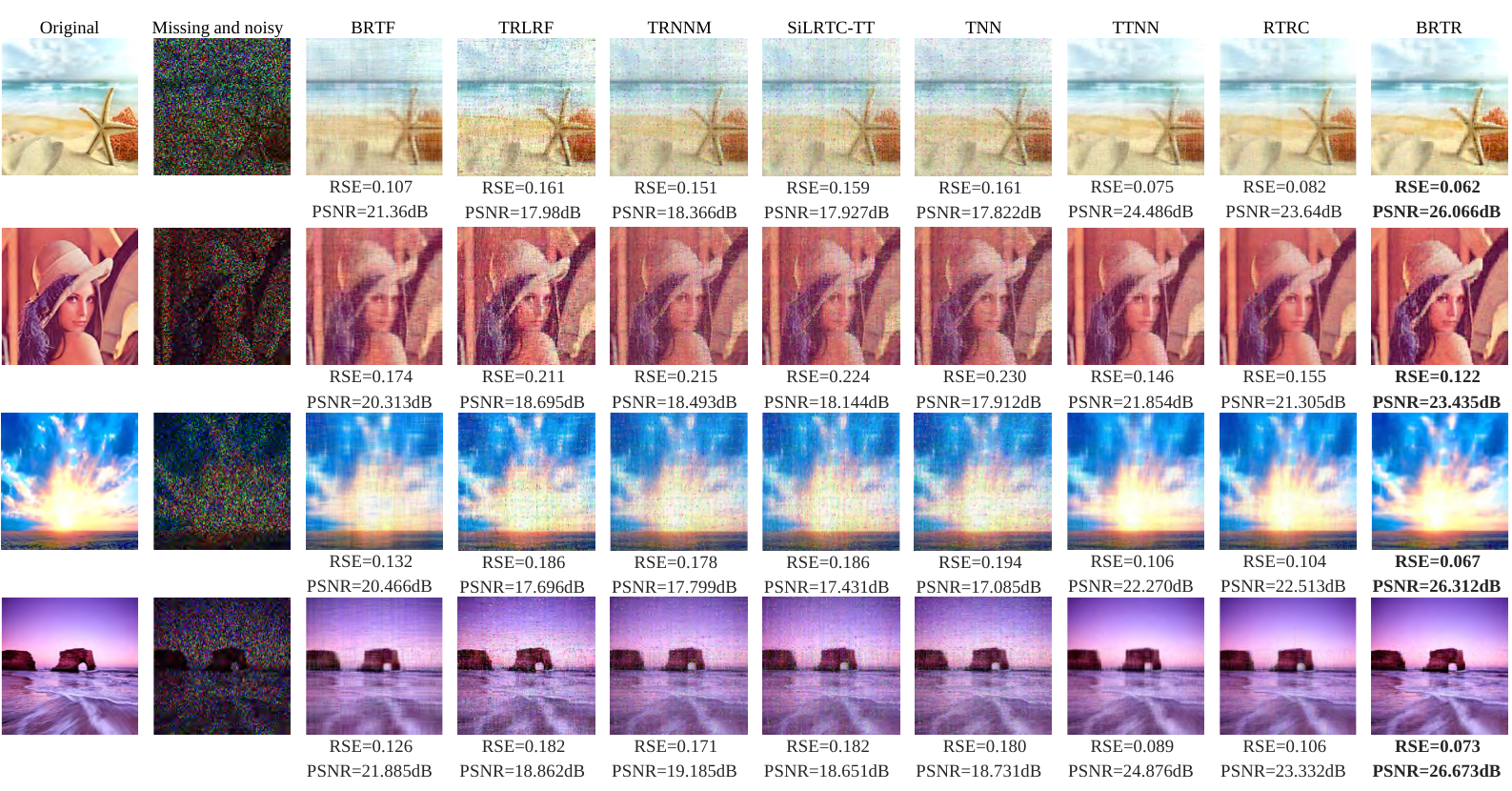}
	\caption{The visualization results of partial color images over BRTF, TRLRF, TRNNM, SiLRTC-TT, TNN, TTNN, RTRC and BRTR methods with MR = 70\% and SR = 10\% in the case of random missing.}
	\label{fig_color_visualization} 
\end{figure*}
\begin{figure*}[h]  
	\centering  
	\includegraphics[width=0.8\linewidth]{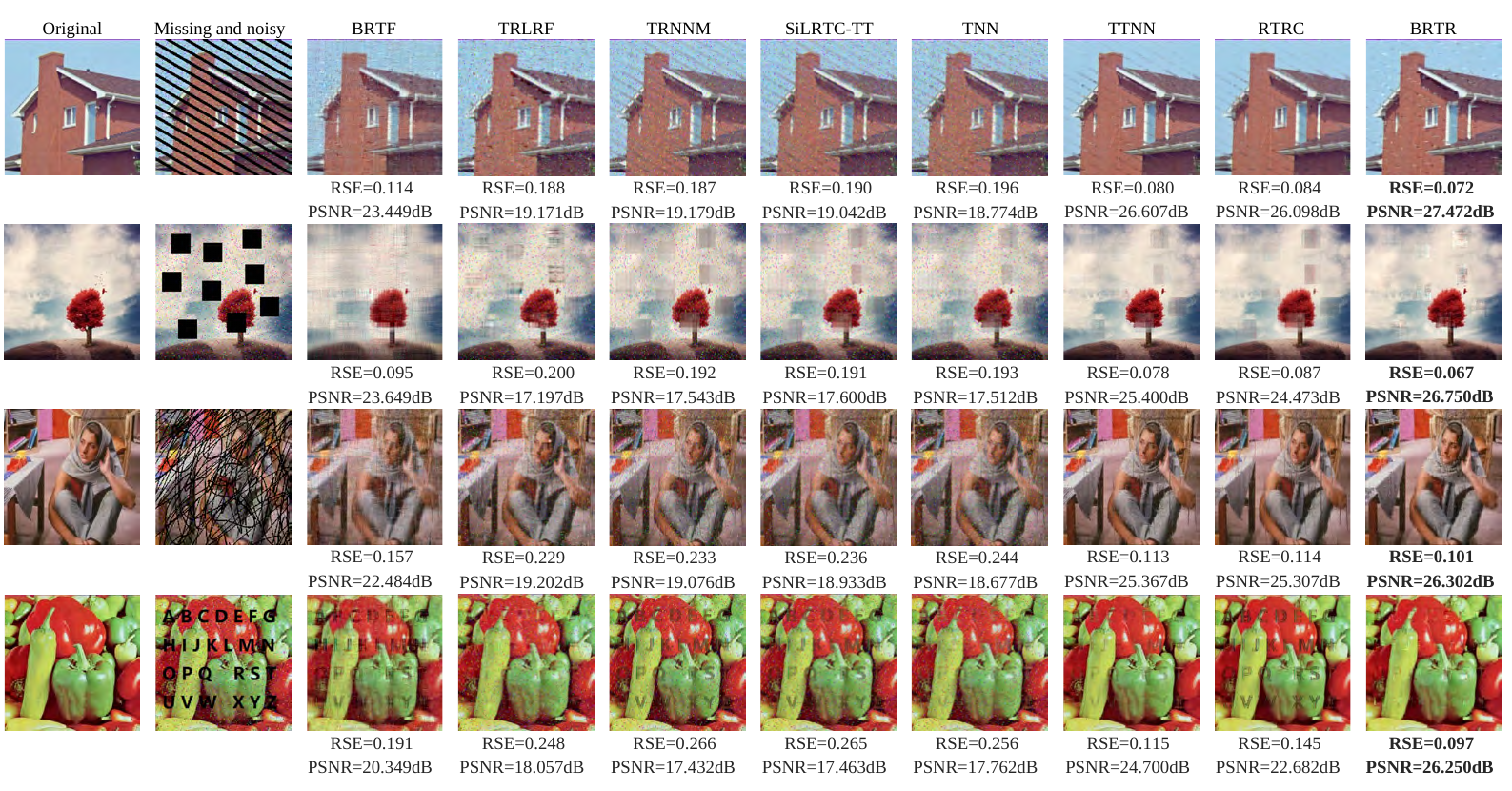}
	\caption{The recovery results on four color images over BRTF, TRLRF, TRNNM, SiLRTC-TT, TNN, TTNN, RTRC and BRTR methods with MR = 70\% and SR = 10\% under different non-random missing pattern. From top to bottom, they are oblique stripe missing, block missing, scratch missing and letter missing. }
	\label{fig_nonrandom_visualization} 
\end{figure*}
In addition, we evaluate the recovery performance by comparing relative standard error (RSE) and peak signal-to-noise ratio (PSNR). Given a 3-order tensor $\boldsymbol{\mathcal{L}} \in \mathbb{R}^{I_1 \times I_2 \times I_3}$, RSE can be defined by
\begin{equation}
	\operatorname{RSE}= \frac{{\|\hat{\boldsymbol{\mathcal{L}}} - \boldsymbol{\mathcal{L}}\|}_{F}}{{\|\boldsymbol{\mathcal{L}}\|}_{F}},
\end{equation}
PSNR can be defined by
\begin{equation}
	\operatorname{PSNR}=10 \log _{10}\left(\frac{\left\|\boldsymbol{\mathcal{L}}\right\|_{\infty}^{2}}{\frac{1}{I_{1} I_{2} I_{3}}\left\|\hat{\boldsymbol{\mathcal{L}}}-\boldsymbol{\mathcal{L}}\right\|_{F}^{2}}\right),
\end{equation}
where $\hat{\boldsymbol{\mathcal{L}}}$ and $\boldsymbol{\mathcal{L}}$ denote the recovered tensor and original tensor, respectively, ${\|\cdot\|}_{\infty}$ is the infinity norm. A smaller RSE and larger PSNR imply better recovery performance. For BRTR, we reshape the input tensors into 9-order tensors with the size $4 \times 4 \times 4 \times 4 \times 4 \times 4 \times 4 \times 4 \times 3$. Fig. \ref{fig_color_rse} and \ref{fig_color_psnr} report the RSE and PSNR values over eight methods of testing color images. From the experiment results, we can observe that BRTR achieves the best recovery performance compared to other methods. Based RTC methods, including BRTF, TNN, TTNN and BRTR, have a better recovery performance than other methods. In addition, compared with RTRC, BRTR have a smaller RSE and larger PSNR, which indicates that the Bayesian inference approach helps to recover images from missing and noisy images. Fig. \ref{fig_color_visualization} shows the visualization results of partial color images in the case of random missing.

The second experiment is set up to evaluate the recovery performance of various methods in removing noise on incomplete observed images in the case of non-random missing. We set four non-random missing patterns, including oblique stripe missing, block missing, scratch missing and letter missing. Similarly, 10\% pixels from observation are randomly corrupted. Fig. \ref{fig_nonrandom_visualization} shows recovery results on four color images under different non-random missing patterns. From Fig. \ref{fig_nonrandom_visualization}, we can observe that BRTR has better recovery performance than other methods in the case of non-random missing. Overall, our proposed BRTR method effectively accomplishes robust tensor completion problems and achieves SOTA performance in incomplete image denoising tasks.

\begin{figure}[t]  
	\centering  
	\subfigure[Indian Pines dataset]{\includegraphics[width=1\linewidth]{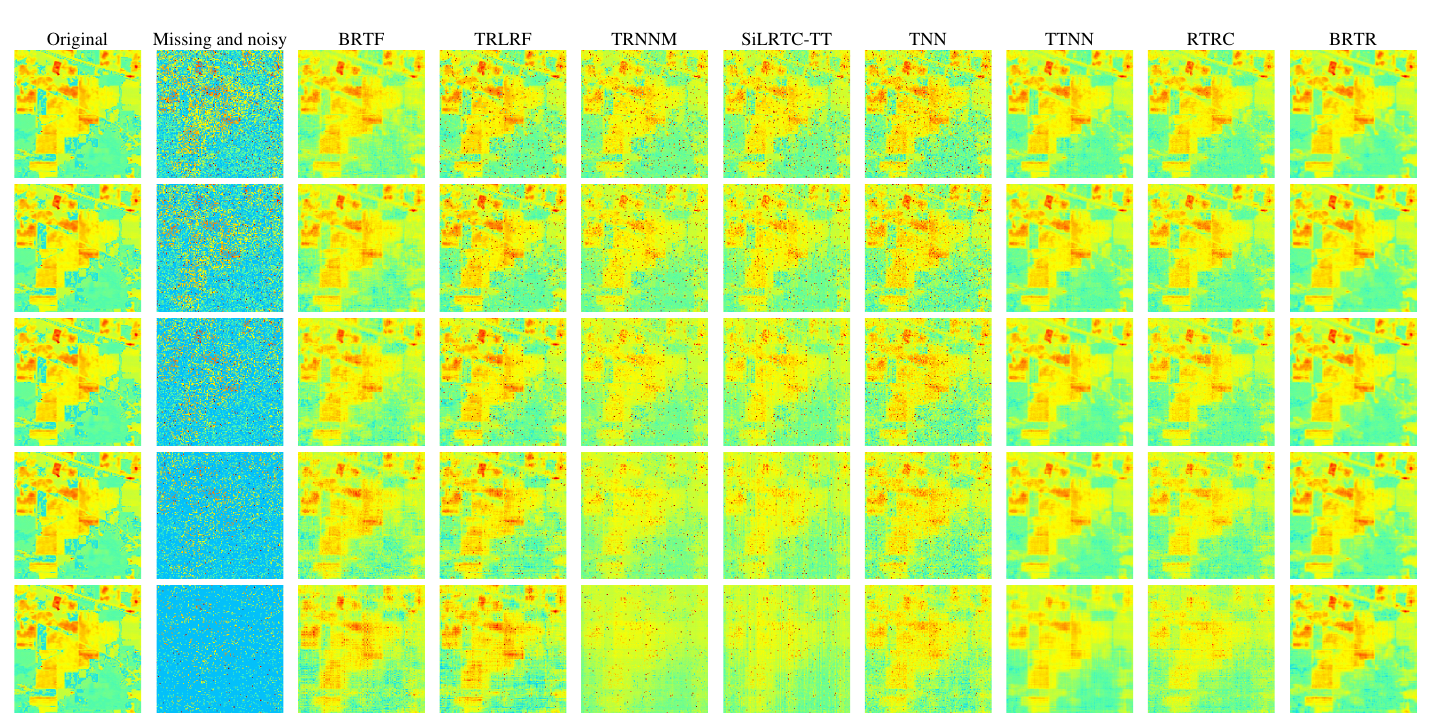}}
	\subfigure[Salinas A dataset]{\includegraphics[width=1\linewidth]{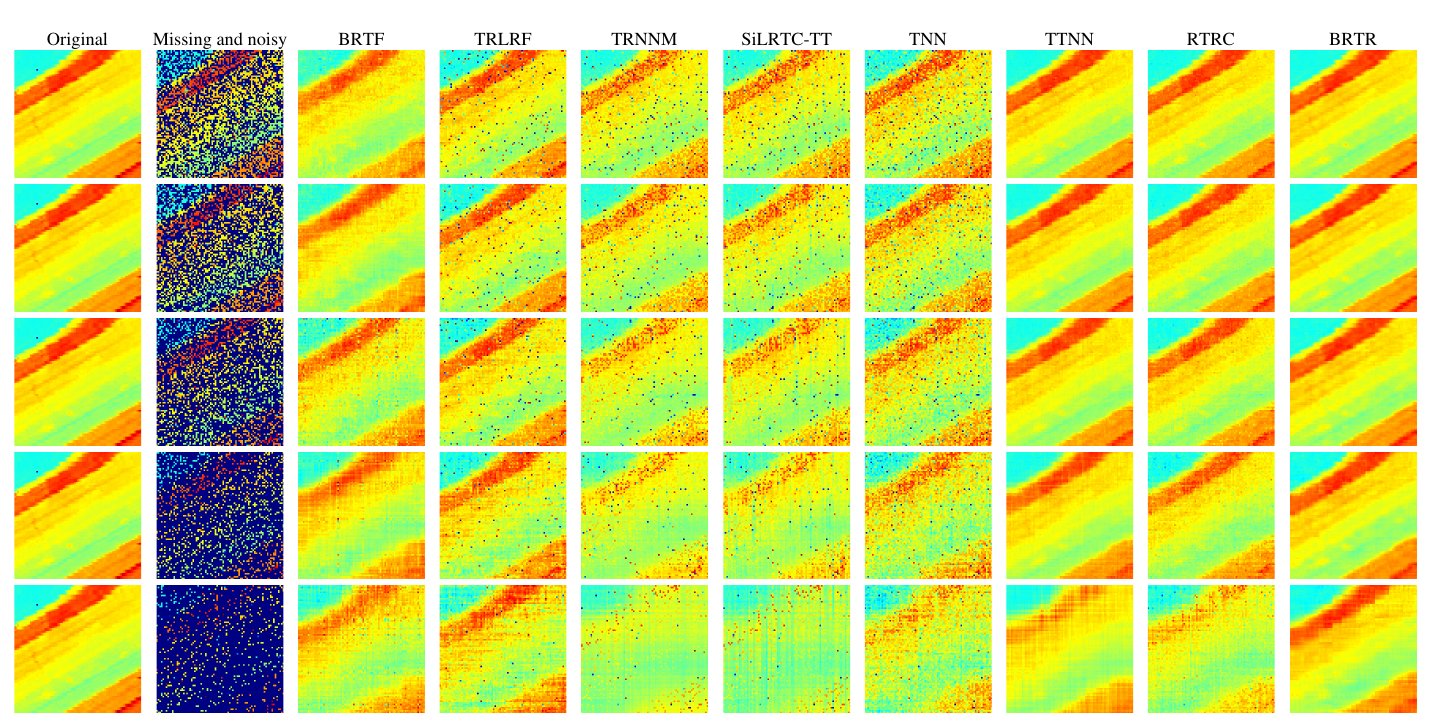}}
	\caption{The visual results over BRTF, TRLRF, TRNNM, SiLRTC-TT, TNN, TTNN, RTRC and BRTR methods under different MR with SR = 10\% on the 21st bound of Indian Pines and Salinas A datasets. (a) Indian Pines dataset. (b) Salinas A dataset. }
	\label{fig_hyper_visualization} 
\end{figure}
\begin{table*}[htb]
	\centering
	\caption{The RSE and PSNR over BRTF, TRLRF, TRNNM, SiLRTC-TT, TNN, TTNN, RTRC and BRTR methods under different MR with SR = 10\% on both hyperspectral datasets. The highest PSNR and lowest RSE are highlighted in $\textbf{bold}$.}
	\renewcommand{\arraystretch}{1.0} 
	\resizebox{\linewidth}{!}{
		\begin{tabular}{m{2cm}<{\centering}|m{0.5cm}<{\centering}|m{1cm}<{\centering}|m{1.2cm}<{\centering}|m{1.2cm}<{\centering}|m{1.2cm}<{\centering}|m{1.2cm}<{\centering}|m{1.2cm}<{\centering}|m{1.2cm}<{\centering}|m{1.2cm}<{\centering}|m{1.2cm}<{\centering}}
			\specialrule{.08em}{0pt}{0pt} 
			Datasets                       & MR                    & Metrics         & BRTF   & TRLRF    & TRNNM  & SiLRTC-TT & TNN    & TTNN   & RTRC   & BRTR   \\
			\specialrule{.08em}{0pt}{0pt} 
			\multirow{10}{*}{Indian Pines} & \multirow{2}{*}{50\%} & RSE$\downarrow$ & 0.135  & 0.252    & 0.267  & 0.272     & 0.291  & 0.089  & 0.089  & \textbf{0.086}  \\
			\cline{3-11}
			&                       & PSNR$\uparrow$  & 27.396 & 21.237   & 20.783 & 20.610    & 19.951 & 31.369 & 30.847 & \textbf{31.639} \\
			\cline{2-11}
			& \multirow{2}{*}{60\%} & RSE$\downarrow$ & 0.146  & 0.236    & 0.258  & 0.264     & 0.285  & 0.098  & 0.102  & \textbf{0.088}  \\
			\cline{3-11}
			&                       & PSNR$\uparrow$  & 26.836 & 21.806   & 21.120 & 20.892    & 20.151 & 30.963 & 29.605 & \textbf{31.453} \\
			\cline{2-11}
			& \multirow{2}{*}{70\%} & RSE$\downarrow$ & 0.156  & 0.220    & 0.251  & 0.258     & 0.277  & 0.109  & 0.118  & \textbf{0.090}  \\
			\cline{3-11}
			&                       & PSNR$\uparrow$  & 26.328 & 22.495   & 21.442 & 21.168    & 20.459 & 30.221 & 28.246 & \textbf{31.123} \\
			\cline{2-11}
			& \multirow{2}{*}{80\%} & RSE$\downarrow$ & 0.171  & 0.207    & 0.254  & 0.261     & 0.271  & 0.129  & 0.146  & \textbf{0.098}  \\
			\cline{3-11}
			&                       & PSNR$\uparrow$  & 25.485 & 23.102   & 21.443 & 21.150    & 20.692 & 28.668 & 26.400 & \textbf{30.265} \\
			\cline{2-11}
			& \multirow{2}{*}{90\%} & RSE$\downarrow$ & 0.211  & 0.220    & 0.284  & 0.297     & 0.274  & 0.176  & 0.208  & \textbf{0.116}  \\
			\cline{3-11}
			&                       & PSNR$\uparrow$  & 23.886 & 22.825   & 20.490 & 19.977    & 20.771 & 25.542 & 23.438 & \textbf{28.658} \\
			\specialrule{.08em}{0pt}{0pt} 
			\multirow{10}{*}{Salinas A}    & \multirow{2}{*}{50\%} & RSE$\downarrow$ & 0.076  & 0.15     & 0.159  & 0.161     & 0.179  & 0.037  & 0.024  & \textbf{0.018}  \\
			\cline{3-11}
			&                       & PSNR$\uparrow$  & 29.237 & 22.113   & 21.582 & 21.452    & 20.606 & 37.370 & 38.114 & \textbf{40.637} \\
			\cline{2-11}
			& \multirow{2}{*}{60\%} & RSE$\downarrow$ & 0.078  & 0.138    & 0.152  & 0.155     & 0.175  & 0.046  & 0.030  & \textbf{0.020}  \\
			\cline{3-11}
			&                       & PSNR$\uparrow$  & 29.151 & 22.796   & 21.995 & 21.809    & 20.829 & 36.516 & 36.375 & \textbf{39.907} \\
			\cline{2-11}
			& \multirow{2}{*}{70\%} & RSE$\downarrow$ & 0.081  & 0.133    & 0.148  & 0.152     & 0.174  & 0.061  & 0.042  & \textbf{0.024}  \\
			\cline{3-11}
			&                       & PSNR$\uparrow$  & 28.370 & 23.436   & 22.304 & 22.062    & 20.977 & 34.832 & 33.720 & \textbf{38.253} \\
			\cline{2-11}
			& \multirow{2}{*}{80\%} & RSE$\downarrow$ & 0.092  & 0.131    & 0.153  & 0.157     & 0.178  & 0.092  & 0.062  & \textbf{0.030}  \\
			\cline{3-11}
			&                       & PSNR$\uparrow$  & 27.344 & 23.786   & 22.275 & 21.977    & 20.992 & 32.178 & 30.395 & \textbf{36.473} \\
			\cline{2-11}
			& \multirow{2}{*}{90\%} & RSE$\downarrow$ & 0.113  & 0.141 & 0.185  & 0.212     & 0.194  & 0.152  & 0.115  & \textbf{0.046}  \\
			\cline{3-11}
			&                       & PSNR$\uparrow$  & 25.417 & 23.161 & 21.072 & 19.883    & 20.682 & 26.331 & 25.361 & \textbf{33.128} \\
			\specialrule{.08em}{0pt}{0pt} 
	\end{tabular}}
	\label{tab_hys_data}
\end{table*}
\begin{table*}[htb]
	\centering
	\caption{The RSE and PSNR over BRTF, OITNN, RTRC and BRTR methods under different SR with MR =70\% for both hyperspectral datasets. The highest PSNR and lowest RSE are highlighted in $\textbf{bold}$.}
	\renewcommand{\arraystretch}{1.0} 
	\resizebox{\linewidth}{!}{
		\begin{tabular}{m{2cm}<{\centering}|m{0.5cm}<{\centering}|m{1cm}<{\centering}|m{1.2cm}<{\centering}|m{1.2cm}<{\centering}|m{1.2cm}<{\centering}|m{1.2cm}<{\centering}|m{1.2cm}<{\centering}|m{1.2cm}<{\centering}|m{1.2cm}<{\centering}|m{1.2cm}<{\centering}}
			\specialrule{.08em}{0pt}{0pt} 
			Datasets                 & SR                    & Metrics         & BRTF   & TRLRF  & TRNNM  & SiLRTC-TT & TNN    & TTNN   & RTRC   & BRTR   \\
			\specialrule{.08em}{0pt}{0pt} 
			\multirow{10}{*}{YaleB1} & \multirow{2}{*}{10\%} & RSE$\downarrow$ & 0.110  & 0.198  & 0.210  & 0.213     & 0.234  & 0.065  & 0.071  & \textbf{0.058}  \\
			\cline{3-11}
			&                       & PSNR$\uparrow$  & 26.746 & 21.159 & 20.664 & 20.538    & 19.711 & 31.092 & 30.437 & \textbf{31.920} \\
			\cline{2-11}
			& \multirow{2}{*}{15\%} & RSE$\downarrow$ & 0.134  & 0.240  & 0.255  & 0.258     & 0.283  & 0.089  & 0.093  & \textbf{0.058}  \\
			\cline{3-11}
			&                       & PSNR$\uparrow$  & 25.029 & 19.484 & 18.971 & 18.878    & 18.055 & 28.292 & 27.941 & \textbf{31.897} \\
			\cline{2-11}
			& \multirow{2}{*}{20\%} & RSE$\downarrow$ & 0.157  & 0.278  & 0.294  & 0.296     & 0.325  & 0.129  & 0.130  & \textbf{0.073}  \\
			\cline{3-11}
			&                       & PSNR$\uparrow$  & 23.600 & 18.223 & 17.743 & 17.676    & 16.852 & 24.940 & 24.950 & \textbf{29.861} \\
			\cline{2-11}
			& \multirow{2}{*}{25\%} & RSE$\downarrow$ & 0.179  & 0.310  & 0.327  & 0.329     & 0.361  & 0.185  & 0.180  & \textbf{0.127}  \\
			\cline{3-11}
			&                       & PSNR$\uparrow$  & 22.511 & 17.264 & 16.805 & 16.753    & 15.932 & 21.784 & 22.031 & \textbf{25.136} \\
			\cline{2-11}
			& \multirow{2}{*}{30\%} & RSE$\downarrow$ & 0.198  & 0.341  & 0.358  & 0.359     & 0.394  & 0.245  & 0.234  & \textbf{0.179}  \\
			\cline{3-11}
			&                       & PSNR$\uparrow$  & 21.544 & 16.448 & 16.032 & 15.993    & 15.171 & 19.323 & 19.727 & \textbf{22.232} \\
			\specialrule{.08em}{0pt}{0pt} 
			\multirow{10}{*}{YaleB2} & \multirow{2}{*}{10\%} & RSE$\downarrow$ & 0.104  & 0.195  & 0.204  & 0.207     & 0.226  & 0.067  & 0.072  & \textbf{0.057}  \\
			\cline{3-11}
			&                       & PSNR$\uparrow$  & 26.815 & 21.090 & 20.593 & 20.445    & 19.696 & 30.521 & 29.988 & \textbf{31.690} \\
			\cline{2-11}
			& \multirow{2}{*}{15\%} & RSE$\downarrow$ & 0.126  & 0.234  & 0.247  & 0.250     & 0.273  & 0.091  & 0.095  & \textbf{0.057}  \\
			\cline{3-11}
			&                       & PSNR$\uparrow$  & 25.109 & 19.440 & 18.907 & 18.800    & 18.043 & 27.739 & 27.412 & \textbf{31.724} \\
			\cline{2-11}
			& \multirow{2}{*}{20\%} & RSE$\downarrow$ & 0.148  & 0.266  & 0.284  & 0.287     & 0.313  & 0.132  & 0.134  & \textbf{0.071}  \\
			\cline{3-11}
			&                       & PSNR$\uparrow$  & 23.711 & 18.276 & 17.690 & 17.613    & 16.848 & 24.460 & 24.356 & \textbf{29.891} \\
			\cline{2-11}
			& \multirow{2}{*}{25\%} & RSE$\downarrow$ & 0.171  & 0.300  & 0.317  & 0.319     & 0.348  & 0.187  & 0.184  & \textbf{0.124}  \\
			\cline{3-11}
			&                       & PSNR$\uparrow$  & 22.493 & 17.248 & 16.750 & 16.692    & 15.932 & 21.402 & 21.508 & \textbf{25.043} \\
			\cline{2-11}
			& \multirow{2}{*}{30\%} & RSE$\downarrow$ & 0.191  & 0.329  & 0.346  & 0.348     & 0.380  & 0.243  & 0.236  & \textbf{0.171}  \\
			\cline{3-11}
			&                       & PSNR$\uparrow$  & 21.455 & 16.435 & 15.972 & 15.925    & 15.171 & 19.075 & 19.306 & \textbf{22.249} \\
			\specialrule{.08em}{0pt}{0pt} 
	\end{tabular}}
	\label{tab_facial_data}
\end{table*}
\subsection{Experiments on Hyperspectral Data}
To further evaluate the effectiveness of our proposed BRTR method, we conduct experiments on hyperspectral image processing, which is a frequent application in the remote sensing scenario. In this section, we evaluate the ability of our BRTR method to remove noise under different MR with SR = 10\% on two hyperspectral datasets, including Indian Pines and Salinas A. The Indian Pines dataset has 224 spectral reflectance bands that has a spatial extent of $145 \times 145$ pixels. For the Salinas A dataset, it contains 224 bands with a spatial extent of $83 \times 86$ pixels. For convenience, we select the first 30 bands of two datasets. Thus we can reshape the data into the size of $145 \times 145 \times 30$ and $83 \times 86 \times 30$ respectively. For BRTR, we need to do size cropping for both datasets and reshape them into the size of $3 \times 4 \times 3 \times 4 \times 3 \times 4 \times 3 \times 4 \times 5 \times 6$ and $3 \times 3 \times 3 \times 3 \times 3 \times 3 \times 3 \times 3 \times 5 \times 6$ respectively. Since both datasets have multiple bands, their PSNR values are defined as the mean of the  PSNR of hyperspectral images of all bands. Fig. \ref{fig_hyper_visualization} shows visual results of robust tensor completion over eight methods on the 21st bound of the Indian Pines dataset and Salinas A datasets and TABLE \ref{tab_hys_data} gives detailed experiment results for both hyperspectral datasets. From TABLE \ref{tab_hys_data}, we can observe that: (1) As MR increases, all eight methods suffer some degree of performance degradation, but the impact of BRTR is relatively small compared to the other methods. (2) At a higher MR, BRTR still can recover images from the missing and noisy observation while other methods fail to do it. The visual results in Fig. \ref{fig_hyper_visualization} can show the superior performance of our BRTR method. Overall, our proposed BRTR method achieves the best performance under different MR, which indicates its stronger recovery capability. 

\subsection{Experiments on Facial Images}
\begin{figure}[t]  
	\centering  
	\subfigure[YaleB1]{\includegraphics[width=1\linewidth]{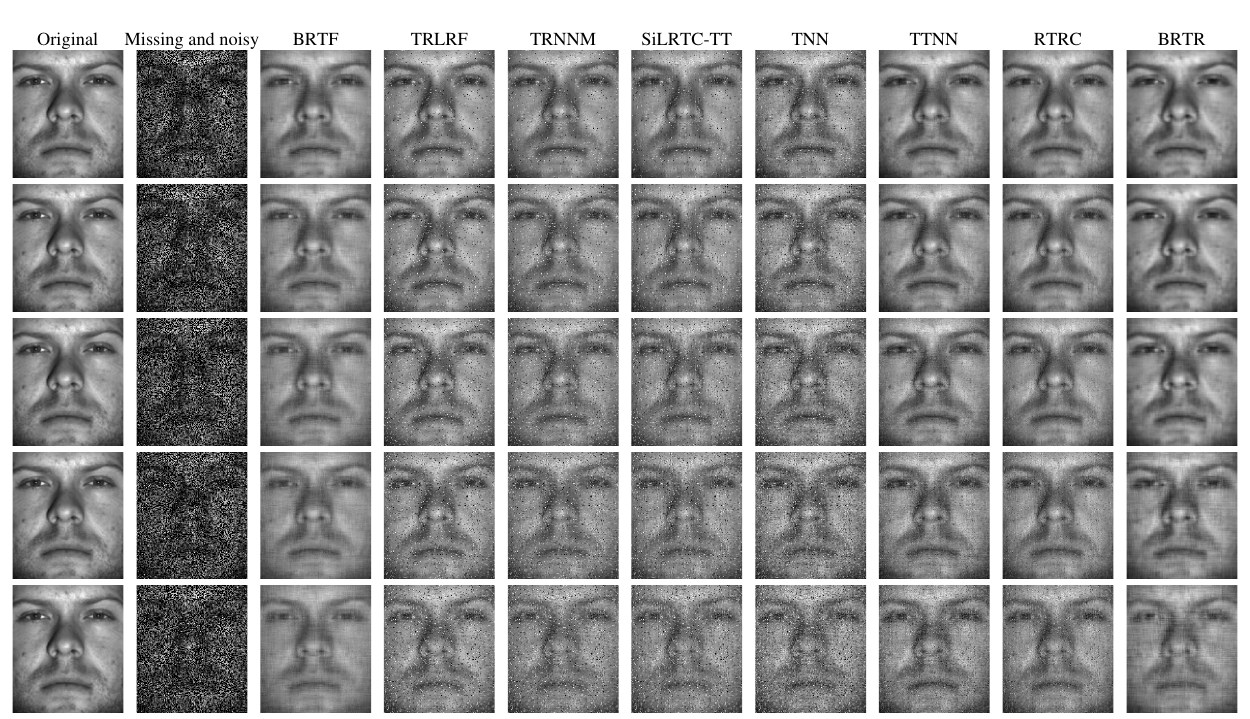}}
	\subfigure[YaleB2]{\includegraphics[width=1\linewidth]{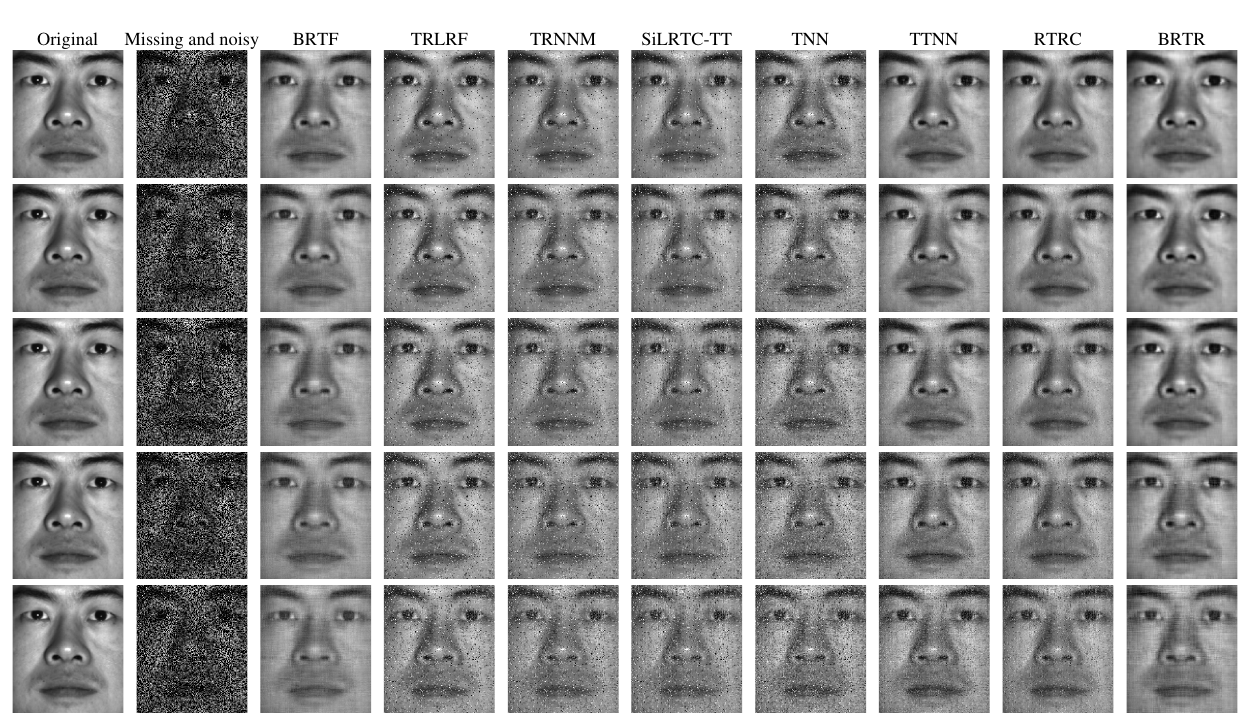}}
	\caption{The visual results over BRTF, TRLRF, TRNNM, SiLRTC-TT, TNN, TTNN, RTRC and BRTR methods under different SR with MR = 70\% on the 1st facial images of YaleB1 and YaleB2 datasets. (a) YaleB1 dataset. (b) YaleB2 dataset. }
	\label{fig_facial_image} 
\end{figure}
\begin{figure*}[t]  
	\centering  
	\includegraphics[width=1\linewidth]{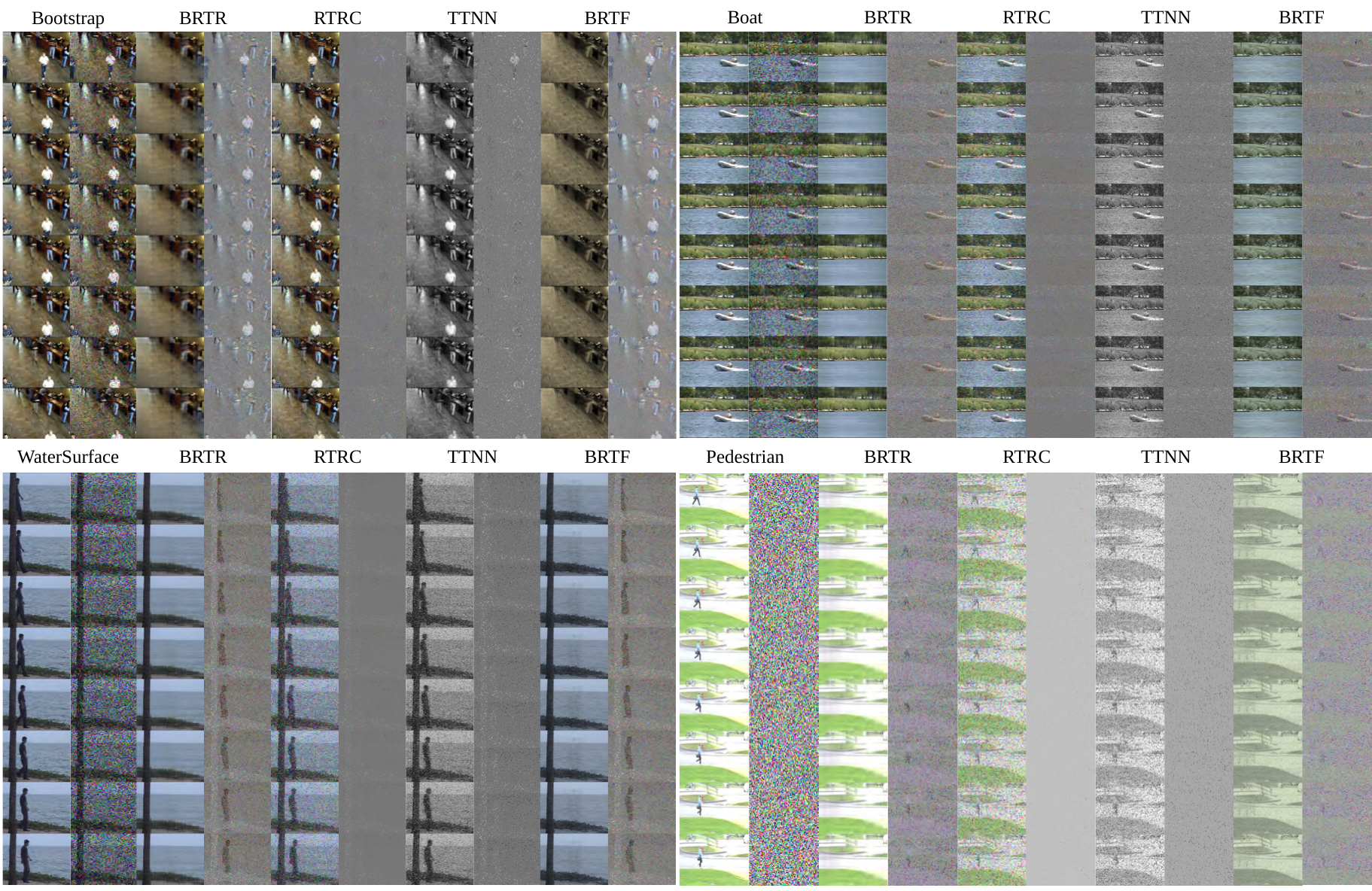}
	\caption{The visual results for the background modeling tasks over BRTR, RTRC, TTNN and BRTF methods on four video datasets with different SR and MR. We select frames 14th to 21st of the WaterSurface dataset and the last 8 frames of the other three datasets for visualization. Each method has two recovered results. The left one is the background and the right one is the separated moving object.}
	\label{fig_video} 
\end{figure*}
In this section, we evaluate the ability of our BRTR method to remove noise under different SR with MR = 70\% on the YaleB face dataset. The YaleB face dataset consists of 16128 multi-pose, multi-illumination images of 20 individuals, specifically including 9 poses and 64 illumination variations. For convenience, we choose the 1st and 2nd subjects and extract their 1st to 12th, 30th to 34th and 36th to 50th frames as YaleB1 and YabeB2 datasets respectively. Each facial image has a size of $192 \times 168$ pixels. Thus we can obtain an input tensor with the size of $192 \times 168 \times 32$. For BRTR, we reshape the input tensors into 9-order tensors with the size of $3 \times 4 \times4\times4\times4\times6\times7\times4\times8$. Fig. \ref{fig_facial_image} shows the visual results on the first facial images of YaleB1 and YaleB2 datasets and TABLE \ref{tab_facial_data} reports the whole experiment results. From the experiment results, we can observe that as SR increases, all methods are increasingly difficult to remove noise. Both TTNN and RTRC have better recovery performance at low SR and poor recovery at high SR. BRTF has a stronger ability to remove noise, and it has better recovery performance than TTNN and RTRC when higher SR. Our BRTR method has the best recovery results in all cases. Through the experimental results of BRTF and BRTR, we can conclude that the Bayesian method can indeed remove the noise better, especially in high SR. In general, our BRTR method has better results in removing noise under different SR from missing and noisy data, which indicates its stronger ability to remove noise for RTC tasks. 

\subsection{Experiments on Video Background Modeling}
This experiment evaluates BRTR in the real surveillance video datasets, including WaterSurface, Boat, Pedestrian and Bootstraps videos. The background of surveillance video datasets can be viewed as a low-rank tensor while the moving object can be viewed as a sparse tensor. Hence, we can treat them as robust tensor completion problems. The WaterSurface video comes from I2R dataset with the size of $128 \times 160 \times 3 \times 300$ and we take its 181st to 229th frames. The Boat and Pedestrian videos come from CDnet dataset with the size of $120 \times 160 \times 3 \times 49$. The Bootstraps video comes from wallflower paper with the size of $120 \times 160 \times 3 \times 49$. Since TTNN can only handle 3-order tensor data, so we remove RGB channels of four videos to ensure that it can get better recovery results. For video background modeling tasks, we simply reshape the input tensors into 5-order tensors where the last dimension of four videos is reshaped as $7 \times 7$. In this experiment, we consider the background modeling results with different SR and MR. (1) Setting1: Bootstrap with SR = 10\% and MR = 10\%; Setting2: Boat with SR = 20\% and MR = 20\%;  Setting3: WaterSurface with SR = 30\% and MR = 30\%; Setting4: Pedestrian with SR = 40\% and MR = 40\%. Fig. \ref{fig_video} visualizes the background modeling results over BRTF, RTRC, RTC, TNN and BRTR methods on four video datasets with different SR and MR. From Fig. \ref{fig_video}, we can observe that both RTRC and TTNN can only separate partial noise and fail to eliminate the moving object in missing and noisy videos. Although BRTF is also able to separate the moving objects, it fails to recover the background and loses a lot of color information. Our BRTR method can successfully separate low-rank backgrounds and sparse noise points in missing and noisy videos. It outperforms all other methods, which also proves its effectiveness in video background modeling. Compared with RTRC and TNN, BRTF and BRTR have better recovery performance for video tasks, which also indicates the advantage of Bayesian approaches. In summary, we can conclude that our proposed BRTR method has a stronger ability to remove noise on incomplete data for different tasks.

\section{Conclusion}
\label{concl}
In this paper, we propose a Bayesian Robust Tensor Ring (BRTR) decomposition method, which can effectively recover target data from missing and noisy observations. BRTR not only automatically learns the TR rank from observations without manual parameter adjustment, but also achieves the balance between the low-rank component and the sparse component, thus enabling better recovery performance. Extensive experiments on synthetic data, colour images, hyperspectral data, video data and face images demonstrate the superiority of BRTR over other SOTA methods. Since the algorithm of BRTR becomes computationally expensive for higher order tensors, we are highly interested in how to reduce its complexity and computation time in the future.

\bibliographystyle{IEEEtran}
\bibliography{BRTR}

\begin{thebibliography}{10}
\providecommand{\url}[1]{#1}
\csname url@samestyle\endcsname
\providecommand{\newblock}{\relax}
\providecommand{\bibinfo}[2]{#2}
\providecommand{\BIBentrySTDinterwordspacing}{\spaceskip=0pt\relax}
\providecommand{\BIBentryALTinterwordstretchfactor}{4}
\providecommand{\BIBentryALTinterwordspacing}{\spaceskip=\fontdimen2\font plus
\BIBentryALTinterwordstretchfactor\fontdimen3\font minus
  \fontdimen4\font\relax}
\providecommand{\BIBforeignlanguage}[2]{{%
\expandafter\ifx\csname l@#1\endcsname\relax
\typeout{** WARNING: IEEEtran.bst: No hyphenation pattern has been}%
\typeout{** loaded for the language `#1'. Using the pattern for}%
\typeout{** the default language instead.}%
\else
\language=\csname l@#1\endcsname
\fi
#2}}
\providecommand{\BIBdecl}{\relax}
\BIBdecl

\bibitem{qiu2021canonical}
Y.~Qiu, G.~Zhou, Y.~Zhang, and A.~Cichocki, ``Canonical polyadic decomposition
  (cpd) of big tensors with low multilinear rank,'' \emph{Multimedia Tools and
  Applications}, vol.~80, no.~15, pp. 22\,987--23\,007, 2021.

\bibitem{zhou2015efficient}
G.~Zhou, A.~Cichocki, Q.~Zhao, and S.~Xie, ``Efficient nonnegative tucker
  decompositions: Algorithms and uniqueness,'' \emph{IEEE Transactions on Image
  Processing}, vol.~24, no.~12, pp. 4990--5003, 2015.

\bibitem{huang2022dynamic}
Z.~Huang, G.~Zhou, Y.~Qiu, Y.~Yu, and H.~Dai, ``A dynamic hypergraph
  regularized non-negative tucker decomposition framework for multiway data
  analysis,'' \emph{International Journal of Machine Learning and Cybernetics},
  pp. 1--20, 2022.

\bibitem{qiu2020generalized}
Y.~Qiu, G.~Zhou, Y.~Wang, Y.~Zhang, and S.~Xie, ``A generalized graph
  regularized non-negative tucker decomposition framework for tensor data
  representation,'' \emph{IEEE transactions on cybernetics}, 2020.

\bibitem{qiu2021approximately}
Y.~Qiu, W.~Sun, Y.~Zhang, X.~Gu, and G.~Zhou, ``Approximately orthogonal
  nonnegative tucker decomposition for flexible multiway clustering,''
  \emph{Science China Technological Sciences}, vol.~64, no.~9, pp. 1872--1880,
  2021.

\bibitem{qiu2021semi}
Y.~Qiu, G.~Zhou, X.~Chen, D.~Zhang, X.~Zhao, and Q.~Zhao, ``Semi-supervised
  non-negative tucker decomposition for tensor data representation,''
  \emph{Science China Technological Sciences}, vol.~64, no.~9, pp. 1881--1892,
  2021.

\bibitem{kolda2009tensor}
T.~G. Kolda and B.~W. Bader, ``Tensor decompositions and applications,''
  \emph{SIAM review}, vol.~51, no.~3, pp. 455--500, 2009.

\bibitem{qiu2022efficient}
Y.~Qiu, G.~Zhou, Z.~Huang, Q.~Zhao, and S.~Xie, ``Efficient tensor robust pca
  under hybrid model of tucker and tensor train,'' \emph{IEEE Signal Processing
  Letters}, vol.~29, pp. 627--631, 2022.

\bibitem{he2017pattern}
H.~He and Y.~Tan, ``Pattern clustering of hysteresis time series with
  multivalued mapping using tensor decomposition,'' \emph{IEEE Transactions on
  Systems, Man, and Cybernetics: Systems}, vol.~48, no.~6, pp. 993--1004, 2017.

\bibitem{sun2021new}
T.~Sun and X.-M. Sun, ``New results on classification modeling of noisy tensor
  datasets: A fuzzy support tensor machine dual model,'' \emph{IEEE
  Transactions on Systems, Man, and Cybernetics: Systems}, vol.~52, no.~8, pp.
  5188--5200, 2021.

\bibitem{huang2021recognition}
Z.~Huang, Y.~Qiu, and W.~Sun, ``Recognition of motor imagery eeg patterns based
  on common feature analysis,'' \emph{Brain-Computer Interfaces}, vol.~8,
  no.~4, pp. 128--136, 2021.

\bibitem{liu2019low}
Y.~Liu, Z.~Long, H.~Huang, and C.~Zhu, ``Low cp rank and tucker rank tensor
  completion for estimating missing components in image data,'' \emph{IEEE
  Transactions on Circuits and Systems for Video Technology}, vol.~30, no.~4,
  pp. 944--954, 2019.

\bibitem{zhang2022effective}
Y.~Zhang, Y.~Wang, Z.~Han, Y.~Tang \emph{et~al.}, ``Effective tensor completion
  via element-wise weighted low-rank tensor train with overlapping ket
  augmentation,'' \emph{IEEE Transactions on Circuits and Systems for Video
  Technology}, 2022.

\bibitem{shao2022tucker}
P.~Shao, D.~Zhang, G.~Yang, J.~Tao, F.~Che, and T.~Liu, ``Tucker
  decomposition-based temporal knowledge graph completion,''
  \emph{Knowledge-Based Systems}, vol. 238, p. 107841, 2022.

\bibitem{kuang2018feature}
H.~Kuang, L.~Chen, L.~L.~H. Chan, R.~C. Cheung, and H.~Yan, ``Feature selection
  based on tensor decomposition and object proposal for night-time multiclass
  vehicle detection,'' \emph{IEEE Transactions on Systems, Man, and
  Cybernetics: Systems}, vol.~49, no.~1, pp. 71--80, 2018.

\bibitem{wu2021latent}
D.~Wu, Y.~He, X.~Luo, and M.~Zhou, ``A latent factor analysis-based approach to
  online sparse streaming feature selection,'' \emph{IEEE Transactions on
  Systems, Man, and Cybernetics: Systems}, vol.~52, no.~11, pp. 6744--6758,
  2021.

\bibitem{nickel2013tensor}
M.~Nickel and V.~Tresp, ``Tensor factorization for multi-relational learning,''
  in \emph{Joint European Conference on Machine Learning and Knowledge
  Discovery in Databases}.\hskip 1em plus 0.5em minus 0.4em\relax Springer,
  2013, pp. 617--621.

\bibitem{ermics2015link}
B.~Ermi{\c{s}}, E.~Acar, and A.~T. Cemgil, ``Link prediction in heterogeneous
  data via generalized coupled tensor factorization,'' \emph{Data Mining and
  Knowledge Discovery}, vol.~29, no.~1, pp. 203--236, 2015.

\bibitem{acar2011scalable}
E.~Acar, D.~M. Dunlavy, T.~G. Kolda, and M.~M{\o}rup, ``Scalable tensor
  factorizations for incomplete data,'' \emph{Chemometrics and Intelligent
  Laboratory Systems}, vol. 106, no.~1, pp. 41--56, 2011.

\bibitem{sorber2013optimization}
L.~Sorber, M.~Van~Barel, and L.~De~Lathauwer, ``Optimization-based algorithms
  for tensor decompositions: Canonical polyadic decomposition, decomposition in
  rank-($l_r$,$l_r$,1) terms, and a new generalization,'' \emph{SIAM Journal on
  Optimization}, vol.~23, no.~2, pp. 695--720, 2013.

\bibitem{de2008tensor}
V.~De~Silva and L.-H. Lim, ``Tensor rank and the ill-posedness of the best
  low-rank approximation problem,'' \emph{SIAM Journal on Matrix Analysis and
  Applications}, vol.~30, no.~3, pp. 1084--1127, 2008.

\bibitem{filipovic2015tucker}
M.~Filipovi{\'c} and A.~Juki{\'c}, ``Tucker factorization with missing data
  with application to low-n-rank tensor completion,'' \emph{Multidimensional
  systems and signal processing}, vol.~26, no.~3, pp. 677--692, 2015.

\bibitem{liu2012tensor}
J.~Liu, P.~Musialski, P.~Wonka, and J.~Ye, ``Tensor completion for estimating
  missing values in visual data,'' \emph{IEEE transactions on pattern analysis
  and machine intelligence}, vol.~35, no.~1, pp. 208--220, 2012.

\bibitem{oh2018scalable}
S.~Oh, N.~Park, S.~Lee, and U.~Kang, ``Scalable tucker factorization for sparse
  tensors-algorithms and discoveries,'' in \emph{2018 IEEE 34th International
  Conference on Data Engineering (ICDE)}.\hskip 1em plus 0.5em minus
  0.4em\relax IEEE, 2018, pp. 1120--1131.

\bibitem{lee2018fast}
D.~Lee, J.~Lee, and H.~Yu, ``Fast tucker factorization for large-scale tensor
  completion,'' in \emph{2018 IEEE International Conference on Data Mining
  (ICDM)}.\hskip 1em plus 0.5em minus 0.4em\relax IEEE, 2018, pp. 1098--1103.

\bibitem{zhang2014novel}
Z.~Zhang, G.~Ely, S.~Aeron, N.~Hao, and M.~Kilmer, ``Novel methods for
  multilinear data completion and de-noising based on tensor-svd,'' in
  \emph{Proceedings of the IEEE conference on computer vision and pattern
  recognition}, 2014, pp. 3842--3849.

\bibitem{lu2019low}
C.~Lu, X.~Peng, and Y.~Wei, ``Low-rank tensor completion with a new tensor
  nuclear norm induced by invertible linear transforms,'' in \emph{Proceedings
  of the IEEE/CVF Conference on Computer Vision and Pattern Recognition}, 2019,
  pp. 5996--6004.

\bibitem{jiang2020multi}
T.-X. Jiang, T.-Z. Huang, X.-L. Zhao, and L.-J. Deng, ``Multi-dimensional
  imaging data recovery via minimizing the partial sum of tubal nuclear norm,''
  \emph{Journal of Computational and Applied Mathematics}, vol. 372, p. 112680,
  2020.

\bibitem{wang2020TNN}
A.~Wang, Z.~Jin, and G.~Tang, ``Robust tensor decomposition via t-svd:
  Near-optimal statistical guarantee and scalable algorithms,'' \emph{Signal
  Processing}, vol. 167, p. 107319, 2020.

\bibitem{wang2020OITNN}
A.~Wang, C.~Li, Z.~Jin, and Q.~Zhao, ``Robust tensor decomposition via
  orientation invariant tubal nuclear norms,'' in \emph{Proceedings of the AAAI
  Conference on Artificial Intelligence}, vol.~34, no.~04, 2020, pp.
  6102--6109.

\bibitem{oseledets2011tensor}
I.~V. Oseledets, ``Tensor-train decomposition,'' \emph{SIAM Journal on
  Scientific Computing}, vol.~33, no.~5, pp. 2295--2317, 2011.

\bibitem{zhao2016tensor}
Q.~Zhao, G.~Zhou, S.~Xie, L.~Zhang, and A.~Cichocki, ``Tensor ring
  decomposition,'' \emph{arXiv preprint arXiv:1606.05535}, 2016.

\bibitem{xu2020learning}
L.~Xu, L.~Cheng, N.~Wong, and Y.-C. Wu, ``Learning tensor train representation
  with automatic rank determination from incomplete noisy data,'' \emph{arXiv
  preprint arXiv:2010.06564}, 2020.

\bibitem{huang2020provable}
H.~Huang, Y.~Liu, J.~Liu, and C.~Zhu, ``Provable tensor ring completion,''
  \emph{Signal Processing}, vol. 171, p. 107486, 2020.

\bibitem{chen2020accommodating}
X.~Chen, G.~Zhou, Y.~Wang, M.~Hou, Q.~Zhao, and S.~Xie, ``Accommodating
  multiple tasks' disparities with distributed knowledge-sharing mechanism,''
  \emph{IEEE Transactions on Cybernetics}, 2020.

\bibitem{yu2021fast}
Y.~Yu, K.~Xie, J.~Yu, Q.~Jiang, and S.~Xie, ``Fast nonnegative tensor ring
  decomposition based on the modulus method and low-rank approximation,''
  \emph{Science China Technological Sciences}, vol.~64, no.~9, pp. 1843--1853,
  2021.

\bibitem{qiu9800181}
Y.~Qiu, G.~Zhou, Q.~Zhao, and S.~Xie, ``Noisy tensor completion via low-rank
  tensor ring,'' \emph{IEEE Transactions on Neural Networks and Learning
  Systems}, pp. 1--15, 2022.

\bibitem{liu2021simulated}
Q.~Liu, X.~Li, H.~Cao, and Y.~Wu, ``From simulated to visual data: A robust
  low-rank tensor completion approach using $l_p$-regression for outlier
  resistance,'' \emph{IEEE Transactions on Circuits and Systems for Video
  Technology}, vol.~32, no.~6, pp. 3462--3474, 2021.

\bibitem{zhang2021multiscale}
H.~Zhang, X.-L. Zhao, T.-X. Jiang, M.~K. Ng, and T.-Z. Huang, ``Multiscale
  feature tensor train rank minimization for multidimensional image recovery,''
  \emph{IEEE Transactions on Cybernetics}, 2021.

\bibitem{yu2020low}
J.~Yu, G.~Zhou, C.~Li, Q.~Zhao, and S.~Xie, ``Low tensor-ring rank completion
  by parallel matrix factorization,'' \emph{IEEE transactions on neural
  networks and learning systems}, 2020.

\bibitem{yu2021robust}
J.~Yu, G.~Zhou, W.~Sun, and S.~Xie, ``Robust to rank selection: Low-rank sparse
  tensor-ring completion,'' \emph{IEEE Transactions on Neural Networks and
  Learning Systems}, 2021.

\bibitem{bengua2017efficient}
J.~A. Bengua, H.~N. Phien, H.~D. Tuan, and M.~N. Do, ``Efficient tensor
  completion for color image and video recovery: Low-rank tensor train,''
  \emph{IEEE Transactions on Image Processing}, vol.~26, no.~5, pp. 2466--2479,
  2017.

\bibitem{ding2019low}
M.~Ding, T.-Z. Huang, T.-Y. Ji, X.-L. Zhao, and J.-H. Yang, ``Low-rank tensor
  completion using matrix factorization based on tensor train rank and total
  variation,'' \emph{Journal of Scientific Computing}, vol.~81, pp. 941--964,
  2019.

\bibitem{yuan2019tensor}
L.~Yuan, C.~Li, D.~Mandic, J.~Cao, and Q.~Zhao, ``Tensor ring decomposition
  with rank minimization on latent space: An efficient approach for tensor
  completion,'' in \emph{Proceedings of the AAAI Conference on Artificial
  Intelligence}, vol.~33, no.~01, 2019, pp. 9151--9158.

\bibitem{yu2019tensor}
J.~Yu, C.~Li, Q.~Zhao, and G.~Zhao, ``Tensor-ring nuclear norm minimization and
  application for visual: Data completion,'' in \emph{ICASSP 2019-2019 IEEE
  international conference on acoustics, speech and signal processing
  (ICASSP)}.\hskip 1em plus 0.5em minus 0.4em\relax IEEE, 2019, pp. 3142--3146.

\bibitem{long2021bayesian}
Z.~Long, C.~Zhu, J.~Liu, and Y.~Liu, ``Bayesian low rank tensor ring for image
  recovery,'' \emph{IEEE Transactions on Image Processing}, vol.~30, pp.
  3568--3580, 2021.

\bibitem{zhao2015bayesian}
Q.~Zhao, G.~Zhou, L.~Zhang, A.~Cichocki, and S.-I. Amari, ``Bayesian robust
  tensor factorization for incomplete multiway data,'' \emph{IEEE transactions
  on neural networks and learning systems}, vol.~27, no.~4, pp. 736--748, 2015.

\bibitem{song2020robust}
G.~Song, M.~K. Ng, and X.~Zhang, ``Robust tensor completion using transformed
  tensor singular value decomposition,'' \emph{Numerical Linear Algebra with
  Applications}, vol.~27, no.~3, p. e2299, 2020.

\bibitem{huang2020robust}
H.~Huang, Y.~Liu, Z.~Long, and C.~Zhu, ``Robust low-rank tensor ring
  completion,'' \emph{IEEE Transactions on Computational Imaging}, vol.~6, pp.
  1117--1126, 2020.

\bibitem{chartrand2007exact}
R.~Chartrand, ``Exact reconstruction of sparse signals via nonconvex
  minimization,'' \emph{IEEE Signal Processing Letters}, vol.~14, no.~10, pp.
  707--710, 2007.

\bibitem{nie2012low}
F.~Nie, H.~Huang, and C.~Ding, ``Low-rank matrix recovery via efficient
  schatten p-norm minimization,'' in \emph{Twenty-sixth AAAI conference on
  artificial intelligence}, 2012.

\bibitem{liu2014exact}
L.~Liu, W.~Huang, and D.-R. Chen, ``Exact minimum rank approximation via
  schatten p-norm minimization,'' \emph{Journal of Computational and Applied
  Mathematics}, vol. 267, pp. 218--227, 2014.

\bibitem{mohan2012iterative}
K.~Mohan and M.~Fazel, ``Iterative reweighted algorithms for matrix rank
  minimization,'' \emph{The Journal of Machine Learning Research}, vol.~13,
  no.~1, pp. 3441--3473, 2012.

\bibitem{cichocki2014era}
A.~Cichocki, ``Era of big data processing: A new approach via tensor networks
  and tensor decompositions,'' \emph{arXiv preprint arXiv:1403.2048}, 2014.

\bibitem{tipping2001sparse}
M.~E. Tipping, ``Sparse bayesian learning and the relevance vector machine,''
  \emph{Journal of machine learning research}, vol.~1, no. Jun, pp. 211--244,
  2001.

\bibitem{wang2017efficient}
W.~Wang, V.~Aggarwal, and S.~Aeron, ``Efficient low rank tensor ring
  completion,'' in \emph{Proceedings of the IEEE International Conference on
  Computer Vision}, 2017, pp. 5697--5705.

\bibitem{wang2019robust}
A.~Wang, X.~Song, X.~Wu, Z.~Lai, and Z.~Jin, ``Robust low-tubal-rank tensor
  completion,'' in \emph{ICASSP 2019-2019 IEEE International Conference on
  Acoustics, Speech and Signal Processing (ICASSP)}.\hskip 1em plus 0.5em minus
  0.4em\relax IEEE, 2019, pp. 3432--3436.

\end{thebibliography}

\end{document}